\PassOptionsToClass{lettersize,journal}{IEEEtran}
\documentclass[lettersize,journal]{IEEEtran}
\usepackage{amsmath,amsfonts}
\usepackage{newtxtext}
\usepackage{algorithmic}
\usepackage{algorithm}
\usepackage{array}
\usepackage[caption=false,font=normalsize,labelfont=sf,textfont=sf]{subfig}
\usepackage{textcomp}
\usepackage{stfloats}
\usepackage{url}
\usepackage{verbatim}
\usepackage{graphicx}
\usepackage{cite}
\usepackage{multirow}
\usepackage{makecell}
\usepackage{gensymb}
\usepackage{xcolor}
\begin{document}

\title{MARVEL: Universal Murray’s Law-informed Vessel Tree Segmentation and Topology Estimation}
% MARVEL: Universal Murray’s Law-informed Vascular Topology Estimation

\author{Yi Zhou, Thiara Sana Ahmed, Jacqueline Chua, Meng Wang, Qinrong Zhang, Alejandro F. Frangi, \IEEEmembership{Fellow, IEEE}, Huazhu Fu, \IEEEmembership{Senior Member, IEEE}, Jun Cheng, \IEEEmembership{Senior Member, IEEE}, Leopold Schmetterer, and Bingyao Tan
\thanks{This work was funded by grants from the National Medical Research Council (OFLCG/004c/2018-00; MOH-000249-00; MOH-000647-00; MOH-001001-00; MOH-001015-00; MOH-000500-00; MOH-000707-00; MOH-001072-06; MOH-001286-00; MOH-001745-00), National Research Foundation Singapore (NRF2019-THE002-0006 and NRF-CRP24-2020-0001), Agency for Science, Technology and Research (M23L7b0021), the Singapore Eye Research Institute \& Nanyang Technological University (SERI-NTU Advanced Ocular Engineering (STANCE) Program), SingHealth Duke-NUS Academic Medicine Research Grant (AM/AIR025/2025), and the SERI-IHPC Joint Lab seeding funding program. 
AFF was supported in part by the Royal Academy of Engineering Chair in Emerging Technologies under Grant CiET1819/19; in part by the Engineering and Physical Sciences Research Council (EPSRC) under Grant EP/Y030494/1 (INSILICO); in part by the British Heart Foundation (BHF) under Grant RE/24/130017 (BHF Manchester Centre of Research Excellence); and in part by the National Institute for Health and Care Research (NIHR) Manchester Biomedical Research Centre under Grant NIHR203308. Corresponding author: B.~Tan (Email: bingyao.tan@duke-nus.edu.sg).}
\thanks{Y.~Zhou and T.~S.~Ahmed are with Singapore Eye Research Institute, Singapore National Eye Centre, Singapore.}
\thanks{J.~Chua is with Singapore Eye Research Institute, Singapore National Eye Centre, Singapore and Ophthalmology \& Visual Sciences Academic Clinical Program, Duke-NUS Medical School, Singapore.}
\thanks{M.~Wang is with Centre for Innovation \& Precision Eye Health, Department of Ophthalmology, Yong Loo Lin School of Medicine, National University of Singapore, Singapore.}
\thanks{Q.~Zhang is with Department of Biomedical Engineering, City University of Hong Kong, Hong Kong, China and Tung Biomedical Sciences Centre, City University of Hong Kong, Hong Kong, China.}
\thanks{A.~F.~Frangi is with Division of Informatics, Imaging, and Data Sciences, Faculty of Biology, Medicine, and Health, School of Health Sciences, University of Manchester, Manchester, United Kingdom; Faculty of Science and Engineering, School of Engineering, University of Manchester, Manchester, United Kingdom; and NIHR Manchester Biomedical Research Centre, Manchester Academic Health Science Centre, Manchester, United Kingdom.}
\thanks{H.~Fu is with Institute of High Performance Computing, Agency for Science, Technology and Research, Singapore.}
\thanks{J.~Cheng is with Institute for Infocomm Research, Agency for Science, Technology and Research, Singapore.}
\thanks{L.~Schmetterer is with Singapore Eye Research Institute, Singapore National Eye Centre, Singapore; SERI-NTU Advanced Ocular Engineering (STANCE) Program, Singapore; Ophthalmology \& Visual Sciences Academic Clinical Program (Eye ACP), Duke-NUS Medical School, Singapore; School of Chemistry, Chemical Engineering and biotechnology, Nanyang Technological University (NTU), Singapore; Centre for Medical Physics and Biomedical Engineering, Medical University of Vienna, Vienna, Austria; Department of Clinical Pharmacology, Medical University of Vienna, Vienna, Austria; and Rothschild Foundation Hospital, Paris, France.}
\thanks{B.~Tan is with Singapore Eye Research Institute, Singapore National Eye Centre, Singapore; SERI-NTU Advanced Ocular Engineering Program, Singapore; and Ophthalmology \& Visual Sciences Academic Clinical Program, Duke-NUS Medical School, Singapore.}
\thanks{We would like to thank Dr. Chi Wei Ong and Dr. Hongying Li from Nanyang Technological University for the fruitful discussion on hemodynamic modeling.}}

\markboth{Journal of \LaTeX\ Class Files,~Vol.~xx, No.~xx, March~2026}%
{Yi Zhou \MakeLowercase{\textit{et al.}}: A Sample Article Using IEEEtran.cls for IEEE Journals}

% % Remember, if you use this you must call \IEEEpubidadjcol in the second column for its text to clear the IEEEpubid mark.

\maketitle

\begin{abstract}
Vascular circulation follows fundamental biophysical principles that optimize mass transport and metabolic energy expenditure, which can be effectively modeled by Murray's law. However, contemporary deep learning methods for vascular segmentation often neglect these biophysical constraints. This leads to physiologically implausible branching and misclassification vascular trees, rendering. These automated segmentation results are unreliable unreliable for downstream clinical tasks such as blood flow simulation or disease quantification. In this paper, we introduce MARVEL (Universal MurrAy's law-infoRmed Vessel sEgmentation and topoLogy estimation), a backbone-agnostic framework that integrates biophysical priors into vascular tree extraction. MARVEL combines per-pixel supervision with explicit radius predictions to enforce local bifurcation constraints derived from an empirical width-exponent mapping. We implement these constraints as differentiable regularizers during training to guide models toward physiologically consistent reconstructions. We evaluate MARVEL on eight public datasets across multiple vascular modalities and segmentation backbones. Results demonstrate MARVEL's superior performance in segmentation accuracy, topological consistency, and physiological plausibility. By converting segmented masks into graph-based hemodynamic simulations, we demonstrate that MARVEL preserves the subtle pathological narrowing and topological connectivity required to distinguish hypertensive from normotensive eyes. Results show that MARVEL significantly improves the classification of hypertension via arteriovenous pressure differences in the eye ($p < 0.001$), outperforming baseline models in both topological consistency and clinical predictive value. The code is available at \url{https://github.com/Zhouyi-Zura/MARVEL}.
\end{abstract}


\section{Introduction}

\begin{figure}[!t]
\centering
\includegraphics[width=0.48\textwidth]{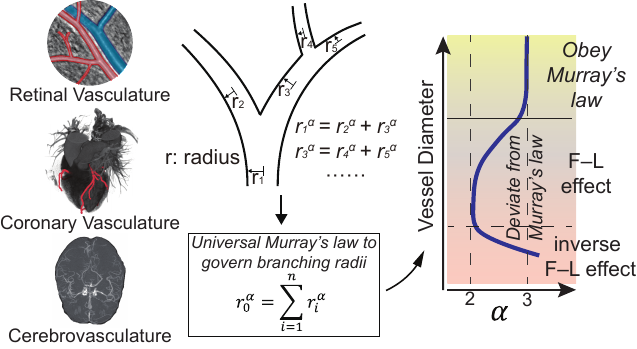}
\caption{Illustration of Murray's law across various vascular modalities. The Murray exponent $\alpha$ depends on the vessel radii $r_i$.}
\label{fig1}
\end{figure}

\IEEEPARstart{A}{ccurate} segmentation of vascular structures is a cornerstone of medical image analysis. It enables the quantitative assessment of vascular biomarkers critical for disease detection, treatment planning, and longitudinal monitoring across multiple imaging modalities, including images of retinal fundus photography, CT/MR angiography, and intravascular imaging. Accurate delineation and reliable vessel reconstruction support downstream applications such as patient-specific hemodynamic modeling, image-guided surgery, and endovascular planning, device sizing and surveillance (e.g., stents and aneurysms), and longitudinal progression studies that track remodeling or treatment response \cite{taylor2009patient}. This is particularly critical in conditions like hypertension, where accurate quantification of peripheral resistance and pressure gradients derived from precise vessel topology can serve as a primary indicator of systemic vascular remodeling \cite{heagerty1993small}.

The primary challenges in vessel analysis involve the accurate handling of multi-scale and multi-orientation tubular structures. These networks contain complex crossings and intricate branching patterns that require more than a simple pixel-level analysis to resolve. Early vessel analysis relied on hand-crafted image-processing pipelines that combined matched filtering, multiscale vessel enhancement, and morphological reconstruction \cite{hoover2000locating,tan2018enhancement}. These traditional tools were later complemented by model-based centerline extraction and profile-fitting techniques \cite{frangi1998multiscale,martinez2007segmentation}. While such approaches effectively leverage vessel geometry priors, they remain vulnerable to low contrast, uneven illumination, complex crossings, small-caliber branches, and pathological deformities, driving the field toward data-driven solutions to improve robustness and generalization.

Deep learning (DL), particularly convolutional neural networks (CNNs), addresses this robustness issue by learning hierarchical representations directly from data. Architectures like U-Net and its volumetric variants significantly improved boundary delineation. Nevertheless, the inherently local receptive fields of CNNs limit their ability to model long-range dependencies, which are essential for resolving complex bifurcations and fine capillaries. To mitigate this, transformer-based frameworks that adopt Vision Transformer (ViT) backbones were introduced to capture global context through self-attention. Hybrid CNN–transformer models further enhance continuity and resilience to noise by balancing local feature extraction with global context modeling.

Despite these architectural expansions, most contemporary methods neglect the biophysical principles governing vascular morphology. By treating vessel segmentation as a texture recognition task rather than a fluid dynamics problem, even state-of-the-art models frequently generate physiologically implausible features, such as abrupt radius changes or disconnected segments. While they handle visual noise better than traditional methods, they lack the intrinsic physical constraints required to guarantee topological validity in hemodynamic modelling/analysis.

The vascular system is a complex, hierarchical network optimized to support life through efficient mass transport and minimal energy expenditure. From the intricate microvasculature of the retina to the high-pressure coronary and pulmonary circuits, this architecture follows universal biophysical principles across diverse organ systems. The most fundamental of these is Murray’s Law \cite{murraylaw}, which prescribes energy-optimal relationships among vessel radii at bifurcations to balance metabolic cost against fluid resistance. As detailed in the recent universal formulation of biological transport \cite{zhou2024universal}, this scaling relationship can be formulated as:
\begin{equation}
r_0^{\alpha} = \sum_{i=1}^{N} r_i^{\alpha},
\label{murrayeq}
\end{equation}
where $r_0$ denotes the parent vessel radius, $r_i$ denotes child branch radius, and $\alpha$ is the Murray exponent. In different vascular systems (Fig.~\ref{fig1}), this universal law captures the biological trade-off between minimal energy expenditure and sufficient perfusion. The right panel of Fig.~\ref{fig1} depicts the relationship between Murray exponent $\alpha$ and vessel diameter, identifying three distinct physiological patterns: large vessels ($> 300 \mu m$) obey Murray's law ($\alpha \approx 3$), intermediate vessels deviate due to the Fahraeus-Lindqvist (F-L) effect ($\alpha$ dips near 2) \cite{fahraeus1931viscosity}, and capillaries show an inverse F-L effect as the $\alpha$ sharply increases \cite{dintenfass1967inversion}.

To address these gaps, we propose \textbf{MARVEL} (Universal \textbf{M}urr\textbf{A}y's law-info\textbf{R}med \textbf{V}essel s\textbf{E}gmentation and topo\textbf{L}ogy estimation). This framework integrates biophysical priors into vascular segmentation in a backbone-agnostic manner. Our approach couples per-pixel supervision with explicit radius-prediction at local bifurcations. Rather than assuming a fixed Murray exponent, MARVEL instead uses a data-driven mapping between vessel width and optimal exponents derived from high-resolution images. By incorporating these constraints as differentiable regularizers, we encourage the network to produce physiologically plausible reconstructions without sacrificing spatial accuracy. Our contributions are fourfold:
\begin{itemize}
\item We develop a flexible framework, \textbf{MARVEL}, that enables the enforcement of Universal Murray's law constraints during training.
\item We construct an empirical mapping between vessel width and Murray exponents, allowing adaptive and radius-specific physiological constraints.
\item We introduce a physics-informed loss that penalizes deviations between the Murray exponents of predicted vessel radii and empirically derived exponents, improving physiological plausibility of predicted vasculature.
\item We perform a comprehensive evaluation on eight public datasets across different vascular modalities. Furthermore, we validate the clinical utility of MARVEL through a hypertension classification task, demonstrating that preserving biophysical integrity enables the calculation of a new hemodynamic biomarker ($\Delta P_{AV}$) that significantly outperform baseline models in disease detection ($p < 0.001$). 
\end{itemize}

\section{Related Works}
\subsection{Automatic Vessel Segmentation Methods}
Vessel segmentation generally falls into two categories: traditional model-based methods and modern data-driven techniques. Early hand-crafted pipelines leveraged multiscale vessel enhancement and matched filtering to amplify tubular structures \cite{frangi1998multiscale,sato1998three}. These methods utilized intensity ridges, centerline traversal, and profile-fitting to localize lumen geometry with high pixel accuracy \cite{aylward2002initialization,lesage2009review}. Graph-based formulations were later introduced to reconstruct branching geometry and maintain topological consistency by representing the vasculature as a set of nodes and edges. While this representation aids in reconstruction and tracking, it requires careful initialization and reliable upstream detection \cite{dashtbozorg2013automatic}.

DL-based models, particularly U-Net and its variants have outperformed traditional filters in boundary delineation. However, standard pixel-wise losses (e.g., Dice, Cross-Entropy) are topologically agnostic; they often yield fragmented networks when processing fine branches or noisy backgrounds \cite{chen2020deep}. Recent studies attempt to address these issues by incorporating topology-aware objectives. Methods like clDice \cite{clDice}, and centerline-boundary variants \cite{cbdice} penalize breaks. Current research employs the integration of multiscale feature learning \cite{hu2022multi}, topology-aware training \cite{zhou2025masked}, and domain adaptation \cite{gharleghi2022towards}. Others reconnect the gaps iteratively in post-processing \cite{mosinska2018beyond}. However, while these methods improve connectivity, they are purely geometric. They enforce only continuity but no other biophysical plausibility, often preserving topologically connected but hemodynamically invalid vessels with erratic radius changes. 

\subsection{Vessel Segmentation Based on Width Estimation}
Integrating radius prediction directly into the segmentation process is for generating radius consistent vessels. Traditional techniques often employ structural analysis to determine vessel dimensions through the analysis of scale, region, and centerline \cite{martinez1999retinal,uslu2019recursive}. Fitting the cross-sectional intensity of vessels enables sub-pixel radius accuracy \cite{lowell2004measurement}. Furthermore, vessel tracking is incorporated to ensure that width estimation remains consistent throughout the vascular network \cite{rothaus2009separation}.

In contrast, modern DL solutions employ learnable modules to directly predict vessel width. Architectures with dilated convolutions, residual blocks, and multiscale fusion enhance the representation of fine-scale details \cite{ma2020retinal}. Other width-aware methods improve vessel caliber perception by generating width maps \cite{chen2022tw} or employing width-based attention to stratify features by scale \cite{ALVARADOCARRILLO2022118313}. However, these methods treat width estimation as a local, pixel-wise regression task, ignoring the relational nature of the vascular tree that parent vessel mathematically constrains the size of its daughter branches.

\subsection{Physics-Informed Methods in Medical Imaging}
Physics-Informed Neural Networks (PINNs) integrate physical laws to govern data-driven models. By minimizing physics-based residuals alongside data-fidelity terms, PINNs ensure that model outputs satisfy governing partial differential equations (PDEs) \cite{raissi2019physics,karniadakis2021physics}. 

In medical imaging, PINNs have been applied to functional tasks including hemodynamic modeling for brain blood flow and myocardial perfusion quantification \cite{sarabian2022physics,van2022physics}. Mechanics-constrained models were used for strain estimation in cardiac imaging \cite{sanchez2018graph}. Additionally, physics priors derived from tissue recovery parameters helped to solve CT/ultrasound inverse image reconstruction \cite{zhu2018image}. Navier-Stokes equations were applied to reconstruct high-fidelity velocity fields in MRI \cite{shone2023deep} and optical imaging in phantoms \cite{cai2021artificial}, while biomechanical assumptions of tissue were used during image registration to detect plausible deformations \cite{hu2018adversarial}.

Despite these advancements, the use of biophysical priors to guide vascular tree extraction remains unexplored. Existing methods solve for flow within a defined geometry instead of using these laws to define the geometry itself. Our work bridges this gap by introducing a tractable mechanism to incorporate the universal formulation of Murray’s law as a differentiable constraint, ensuring that the segmented structure itself adheres to the principles of hemodynamic optimality.

\begin{figure*}[!ht]
\centering
\includegraphics[width=\textwidth]{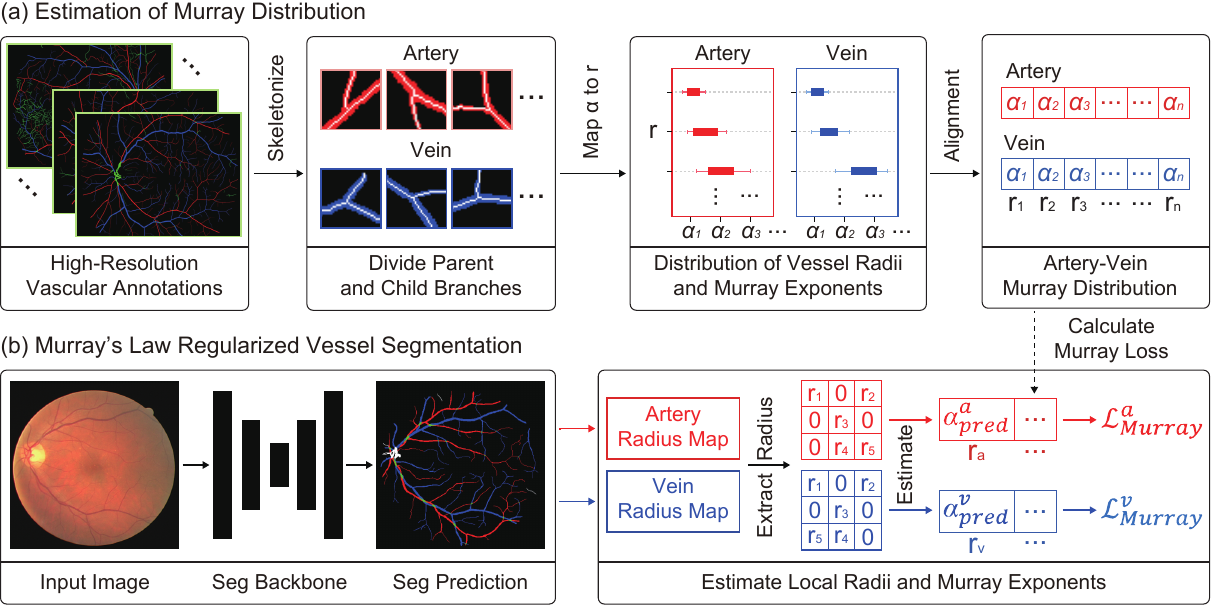}
\caption{Overview of the proposed MARVEL framework. (a) Estimation of the Murray distribution: the relationship between vessel radius and the Murray exponent is established using high-resolution ground truth masks or prior biophysical knowledge. (b) Physics-informed vessel segmentation: the backbone network predicts vessel masks, which are subsequently processed through differentiable skeletonization and junction detection. Local vessel radii at bifurcations are extracted to calculate Murray exponents, facilitating a physics-informed loss function that enforces physiological consistency during model training.}
\label{fig2}
\end{figure*}

\section{Method}
We propose \textbf{MARVEL}, a framework that enforces hemodynamic consistency namely Murray's law, into vascular segmentation. Unlike conventional methods that assume a static branching exponent (typically $\alpha$ = 3), MARVEL dynamically adapts to local geometry via a data-driven width-exponent mapping. This flexibility is critical for handling diverse vascular morphologies where metabolic scaling varies. An overview of the framework is shown in Fig.~\ref{fig2}.

Murray's law governs the preservation of hydraulic conductance at bifurcations. When a parent branch of radius $r_0$ that splits into $N$ child branches of radii $r_i$, their radii should satisfy the Eq.~\ref{murrayeq}, where the component $\alpha$ reflects the specific metabolic demands of the tissue. While $\alpha \approx 3$ \cite{talkington2022dermal} for retinal vessels, it drops to 2.7 for cerebrovasculature \cite{suwa1963estimation} and 2.4 for coronary arteries \cite{taylor2024systematic}. MARVEL avoids the use of rigid assumptions by learning $\alpha$ as a function of vessel width empirically from high-resolution reference data.

\subsection{Data-Driven Discovery of Murray Exponents}
To calibrate these biophysical constraints, we first derive an empirical mapping between vessel width and the Murray exponent $\alpha$. 

\subsubsection{Graph Extraction and Radius Profiling}
We extract topological graphs from high-resolution reference masks (e.g., HRF-AV). The process begins with iterative thinning to produce a unit-width skeleton of the vessel tree. We identify branch points and endpoints as the primary nodes of the graph. To ensure robust radius estimation, we employ a "moat" strategy: dilating the identified nodes with a local spatial kernel and removing these regions from the skeleton. This separation ensures that we measure vessel properties exclusively on linear segments unaffected by bifurcation artifacts. For each remaining segment, we determine the vessel radius using the Euclidean distance transform of the original binary mask. Subsequently, we calculate the final radius by averaging the middle 80\% of the measurements collected along the segment. By discarding the top and bottom 10\%, we effectively filter out noise and junction artifacts.

\subsubsection{Local Murray Exponent Inference}
For every valid bifurcation, we solve the generalized Murray equation numerically. At each node $n_i$, we identify all incident edges and their radii. The edge with the maximum radius is designated as the parent branch ($r_p$), and the remaining edges are child branches $\{r_{c1}, r_{c2}, \ldots\}$. For bifurcations with at least two child branches, we solve for the local Murray exponent $\alpha$ by finding the root of
\begin{equation}
f(\alpha) = r_p^\alpha - \sum_{i=1}^{N} r_{ci}^{\alpha} = 0.
\end{equation}

Solutions yielding non-real or non-positive $\alpha$ values are discarded. This procedure yields a collection of pairs $(w_k, \alpha_k)$, where $w_k = r_p \times s$ denotes the parent vessel width in micrometers. The scale factor $s$ is derived from the imaging modality as detailed in Section~\ref{secdataset}.

By aggregating these solutions across thousands of bifurcations, we construct a discrete lookup table $\mathbf{w}_{GT} = [w_1, w_2, \ldots, w_B]$ to $\boldsymbol{\alpha}_{GT} = [\tilde{\alpha}_1, \tilde{\alpha}_2, \ldots, \tilde{\alpha}_B]$. We sample 2,282 arterial and 2,292 venous branches from the HRF-AV dataset. Following the extraction of radii and exponents in the pixel domain, we convert these measurements to micrometers ($\mu$m) to construct a discrete lookup table with an interval of 1 $\mu$m. The 95\% confidence intervals for the derived Murray exponents $\alpha$ are (2.45, 2.57) for arteries and (2.74, 2.87) for veins, respectively. For coronary artery and cerebrovascular segmentation, we use fixed value 2.39 \cite{taylor2024systematic} and 2.7 \cite{suwa1963estimation} as $\alpha_{GT}$, respectively.

\subsection{Murray’s Law-informed Differentiable Training}
We couple standard segmentation losses with a differentiable Murray loss that penalizes biophysical violations in training.

\subsubsection{Soft Skeletonization and Radius Estimation}
To compute topological losses, we use a soft skeleton scheme with differentiable morphological operations (erosion as $-\text{maxpool}(-p)$ and dilation as $\text{maxpool}(p)$ over a local neighborhood). This process generates a continuous-value skeleton map that preserves gradient flow \cite{clDice}:
\begin{equation}
sk(u) = \frac{ReLU(p_{open}(u) - \bar{p}_{local}(u))}{\max_{u'} ReLU(p_{open}(u') - \bar{p}_{local}(u'))},
\end{equation}
where $\bar{p}_{local}$ is the local average and $u$ denotes spatial coordinates. Skeleton values are normalized between $[0,1]$. Junctions are identified by convolving the soft skeleton with an all-one kernel, followed by a sharp sigmoid activation.

Concurrently, we regress a radius map $rm_{pred}$ supervised by a robust L1 loss:
\begin{equation}
\mathcal L_{rad} = \bigl| rm_{pred} \cdot sk(u) - rm_{gt} \cdot sk(u) \bigr|_1,
\end{equation}
where $rm_{gt}$ is the ground truth radius. This encourages correct scale estimation while de-emphasizing off-center noise.

\subsubsection{The Adaptive Murray Loss}
At every predicted junction, we evaluate Murray’s law in the logarithmic scale. This step effectively mitigates the gradient instability and numerical overflow associated with the power terms of the $\alpha$ parameter. Furthermore, it enables the use of the Log-Sum-Exp (LSE) technique to robustly compute the aggregate daughter capacity while handling the vast dynamic range of vessel radii. This ensures that the summation of child branches strictly adheres to the physical conservation law in a numerically stable manner. The relationship is expressed as:
\begin{equation}
\alpha \log R_p = \log\left(\sum_{i} R_{c,i}^\alpha\right).
\end{equation}

This formulation allows the network to apply strict power laws to the microvasculature while relaxing constraints for larger vessels, mirroring the true biological variation. 
To aggregate child branch contributions, we apply softmax weighting:
\begin{equation}
w_i = \frac{\exp(R_{c,i} / (\tau \cdot \max_j R_{c,j}))}{\sum_{j=1}^{N_n} \exp(R_{c,j} / (\tau \cdot \max_j R_{c,j}))},
\end{equation}
where $\tau=0.1$ controls the weighting temperature, $N_n$ is the number of adjacent elements, and $R_{c,i}$ denotes the $i$-th neighbor radius. 
We evaluate this relationship across all empirically derived exponents $\alpha \in \boldsymbol{\alpha}_{GT}$ to compute per-element, per-exponent errors in log-space. The predicted local Murray exponent $\alpha_{pred}(u)$ is obtained via soft-argmin over the candidate set. Similarly, the target exponent $\alpha_{GT}(u)$ is computed via Gaussian interpolation over the empirical mapping:
\begin{equation}
\alpha_{GT}(u) = \sum_{b=1}^{B} w_b(u) \cdot \tilde{\alpha}_b,
\end{equation}
with Gaussian weights
\begin{equation}
w_b(u) = \frac{\exp\left(-\frac{(\log r_e - \log R_p(u))^2}{2\sigma_{\log}^2}\right)}{\sum_{b'=1}^{B} \exp\left(-\frac{(\log r_{e'} - \log R_p(u))^2}{2\sigma_{\log}^2}\right)},
\end{equation}
where $r_e$ denotes the radius in the empirical mapping, $r_{e'}$ is the indexing variable over all mappings in the normalization term, and $\sigma_{\log}$ is the standard deviation of log-transformed empirical radii.

To improve robustness against outliers and ambiguous regions, we discard the top 10\% of elements with the largest errors before computing the final loss. The element-wise Murray penalty is weighted by the junction probability $j(u)$:
\begin{equation}
\ell_{Murray}(u) = j(u) \cdot (\alpha_{pred}(u) - \alpha_{GT}(u))^2.
\end{equation}

The batch-level Murray loss is:
\begin{equation}
\mathcal L_{Murray} = \frac{\sum_{u} \ell_{Murray}(u)}{\sum_{u} j(u)}.
\end{equation}

We compute $\mathcal L_{Murray}$ independently for the arteries and veins.

\subsubsection{Overall Training Objective}
The learning objective is formulated as a composite loss function that combines standard segmentation objectives with the physics-informed terms to simultaneously optimize for spatial overlap and biophysical consistency:
\begin{equation}
\mathcal L = \mathcal L_{Dice} + \mathcal L_{MSE} + \lambda \mathcal L_{Murray} + \beta \mathcal L_{Rad},
\end{equation}
where $\lambda$ and $\beta$ balance the regularization terms (both are set to 0.1 empirically). The $\mathcal L_{Dice}$ and $\mathcal L_{MSE}$ are calculated between the predicted probability map and the ground truth. We choose them as the pixel-wise fidelity terms because they provide smoother gradients for thin vessels during the early stages of training. This objective encourages the model to preserve spatial accuracy while imposing physiologically meaningful constraints on branching geometry.

\section{Experiments and Results}
\subsection{Datasets and Implementation Details}\label{secdataset}

\begin{table}[!t]
\centering
\caption{Dataset specifications for evaluation. FOV denotes the field of view, and No. represents the number of images. The unit of pixel size for 2D and 3D are $\mu m/pixel$ and $mm/voxel$, respectively.}
\setlength{\tabcolsep}{3pt}
\begin{tabular}{llcccc}
\hline
\textbf{Modality} &\textbf{Dataset} &\textbf{No.} &\textbf{FOV} &\textbf{Pixel Resolution} &\textbf{Pixel Size}\\
\hline
\multirow{6}{*}{\textbf{Fundus}} &RITE \cite{rite} &40 &45$^\circ$ &584$\times$565 &21.1\\
\cline{2-6}
&\multirow{2}{*}{LES-AV \cite{LesAV}}
&21 &30$^\circ$ &1444$\times$1620 &\multirow{2}{*}{5.43}\\
\cline{3-5}
& &1 &45$^\circ$ &1958$\times$2196\\
\cline{2-6}
&\multirow{2}{*}{F-AVSeg \cite{deng2025fundus}}
&79 &45$^\circ$ &1280$\times$1280 &\multirow{2}{*}{9.22}\\
\cline{3-5}
& &21 &45$^\circ$ &2656$\times$1992\\
\cline{2-6}
&HRF-AV \cite{hrf} &45 &60$^\circ$ &2336$\times$3504 &4.81\\
\hline
\multirow{2}{*}{\textbf{CTA}}
&ImageCAS \cite{imageCAS} &1,000 &N/A &233$\times$512$\times$512 &0.70\\
\cline{2-6}
&CCA \cite{cca200} &20 &N/A &576$\times$832$\times$832 &0.60\\
\hline
\multirow{4}{*}{\textbf{TOF-MRA}}
&MIDAS \cite{hilbert2020brave} &20 &N/A &128$\times$448$\times$448 &1.14\\
\cline{2-6}
&\multirow{3}{*}{COSTA \cite{mou2024costa}} &\multirow{3}{*}{423} &\multirow{3}{*}{N/A} &100$\times$256$\times$256 &0.25\\
& & & &to &to\\
& & & &92$\times$1024$\times$1024 &2.15\\
\hline
\end{tabular}
\label{dataset}
\end{table}

We evaluate eight public datasets across three imaging modalities: 2D fundus photography, 3D computed tomography angiography (CTA), and 3D Time-of-Flight Magnetic Resonance Angiography (TOF-MRA). Table~\ref{dataset} details these datasets. To ensure reproducibility and fair comparison, we adopt the official training and test splits for all established benchmarks. 

We train and evaluate the model for each dataset independently. Models are trained end-to-end using the Adam optimizer (learning rate: $10^{-3}$, beta1 = 0.5, beta2 = 0.999) with a batch size of 8 and up to 1000 epochs. Augmentation is restricted to random horizontal flips. To balance memory efficiency and resolution, we train the model on patches. These patches have dimensions of $512\times512$ for RITE, $1024\times1024$ for LES-AV, F-AVSeg, and HRF-AV, and $128\times128\times128$ for ImageCAS, CCA, MIDAS, and COSTA. Testing is conducted on padded patches without overlap and aggregated to final full-image predictions.

A critical prerequisite for physics-informed learning is the standardization of physical units. To ensure that Murray regularization is implemented on a consistent physical scale, we convert pixel/voxel-based measurements into $\mu m$ or $mm$. We use a standard retinal scale of $0.26\ mm$ per degree of visual angle \cite{provis2013adaptation} and the pixel size from each dataset is summarized in Table~\ref{dataset}. 

We use PyTorch to maintain gradient flow throughout the pipeline. Experiments are conducted on a single NVIDIA H100 NVL GPU with 94 GB of memory. The typical training times are roughly one hour for RITE and LES-AV, two hours for F-AVSeg and HRF-AV, ten hours for CCA and MIDAS, and thirty hours for ImageCAS and COSTA.

\subsection{Evaluation Metrics}
We assess the performance of the framework using several complementary metrics. These metrics evaluate classification accuracy, segmentation fidelity, and the integrity of the topological and physiological structures.

\subsubsection{Spatial Segmentation Fidelity}
\begin{itemize}
\item \textbf{Accuracy ($Acc$)}: This metric measures the overall correctness of the vessel classification.
\item \textbf{Hausdorff Distance ($HD$)}: This surface-based metric measures the maximum Euclidean distance between points on the predicted and reference vessel surfaces in volumetric space. It captures the extreme geometric discrepancies in 3D segmentation.
\end{itemize}

\subsubsection{Topological Fidelity}
\begin{itemize}
\item \textbf{Dice coefficient ($Dice$)} and \textbf{centerline Dice ($clDice$)} \cite{clDice}: These are overlap-based metrics for segmentation quality. The $clDice$ specifically focuses on the preservation of thin vascular structures by comparing the intersections of skeletons.
\item \textbf{Connectivity, Area, Length ($CAL$)} \cite{gegundez2011function}: This composite metric is specifically designed for vascular structures. It quantifies fragmentation, overlap, and coincidence between the predicted and reference vessel graphs.
\item \textbf{Betti Number error ($\beta^{err}$)} \cite{hu2019topology} and \textbf{Betti Matching error ($\mu^{err}$)} \cite{stucki2023topologically}: These topological metrics quantify discrepancies in 0-dimensional (connected components) and 1-dimensional (holes or cycles) Betti numbers. Betti Matching additionally accounts for spatial correspondence of these components. We report both 0-d and 1-d errors ($\beta_0^{err},\mu_0^{err}$ and $\beta_1^{err},\mu_1^{err}$).
\end{itemize}

\begin{table*}[!ht]
\centering
\caption{Comprehensive performance comparison on four retinal A/V datasets (mean $\pm$ standard deviation). $\uparrow$ means higher values are better, and $\downarrow$ vice versa. \textbf{Bold} indicates a significant improvement, \textit{italics} indicates at least one non-significant improvement.}
\setlength{\tabcolsep}{7pt}
\begin{tabular}{cl|cccccccc}
\hline
\textbf{Dataset} & \textbf{Method} & \textbf{Acc (\%)$\uparrow$} & \textbf{Dice (\%)$\uparrow$} & \textbf{clDice (\%)$\uparrow$}& \textbf{CAL (\%)$\uparrow$} & $\beta_0^{err}\downarrow$ & $\beta_1^{err}\downarrow$ & $\mu_0^{err}\downarrow$ & $\mu_1^{err}\downarrow$\\
\hline
\multirow{9}{*}{\rotatebox{90}{\textbf{RITE}}}
& TW-GAN \cite{chen2022tw}
& 98.13 $\pm$ 0.22 & 68.61 $\pm$ 0.87 & 81.33 $\pm$ 2.99 & 30.15 $\pm$ 2.34 & 0.346 & 0.728 & 0.482 & 0.351 \\
& CF-Loss \cite{zhou2024cf}
& 98.18 $\pm$ 0.21 & 68.95 $\pm$ 0.92 & 81.20 $\pm$ 3.08 & 31.62 $\pm$ 2.82 & 0.305 & 0.655 & 0.476 & 0.345 \\
& RRWNet \cite{morano2024rrwnet}
& 98.15 $\pm$ 0.19 & 69.24 $\pm$ 0.94 & 81.44 $\pm$ 2.92 & 31.59 $\pm$ 2.98 & 0.310 & 0.682 & 0.466 & 0.352 \\
& clDice \cite{clDice}
& 98.08 $\pm$ 0.19 & 67.67 $\pm$ 0.88 & 81.06 $\pm$ 3.02 & 30.45 $\pm$ 2.46 & 0.331 & 0.795 & 0.494 & 0.353 \\
& TopoLoss \cite{hu2019topology}
& 98.09 $\pm$ 0.18 & 68.39 $\pm$ 0.87 & 80.90 $\pm$ 3.04 & 31.36 $\pm$ 2.43 & 0.308 & 0.722 & 0.485 & 0.379 \\
& BettiLoss \cite{stucki2023topologically}
& 97.98 $\pm$ 0.19 & 67.23 $\pm$ 0.89 & 81.26 $\pm$ 3.10 & 31.09 $\pm$ 2.51 & 0.301 & 0.679 & 0.475 & 0.349 \\
& Gupta et al. \cite{gupta2024topology}
& 98.08 $\pm$ 0.18 & 69.07 $\pm$ 0.86 & 81.17 $\pm$ 2.98 & 32.27 $\pm$ 2.39 & 0.298 & 0.588 & 0.469 & 0.34 \\
& MaskVSC \cite{zhou2025masked}
& 98.13 $\pm$ 0.17 & 70.22 $\pm$ 0.84 & 81.30 $\pm$ 2.94 & 33.01 $\pm$ 2.37 & 0.294 & 0.577 & 0.463 & 0.338 \\
& \textbf{Proposed}
& \textbf{98.28 $\pm$ 0.18} & \textbf{71.12 $\pm$ 0.92} & \textbf{81.78 $\pm$ 2.65} & \textbf{36.99 $\pm$ 3.67} & \textbf{0.277} & \textbf{0.505} & \textbf{0.446} & \textbf{0.317} \\
\hline
\multirow{9}{*}{\rotatebox{90}{\textbf{LES-AV}}}
& TW-GAN \cite{chen2022tw}
& 98.82 $\pm$ 0.23 & 71.08 $\pm$ 1.05 & 79.02 $\pm$ 7.55 & 24.62 $\pm$ 6.80 & 2.01 & 0.32 & 0.255 & 0.237 \\
& CF-Loss \cite{zhou2024cf}
& 98.83 $\pm$ 0.21 & 71.43 $\pm$ 1.00 & 79.50 $\pm$ 7.47 & 24.77 $\pm$ 6.45 & 1.89 & 0.313 & 0.243 & 0.211 \\
& RRWNet \cite{morano2024rrwnet}
& 98.81 $\pm$ 0.21 & 71.36 $\pm$ 1.04 & 79.23 $\pm$ 7.15 & 24.82 $\pm$ 6.51 & 1.98 & 0.335 & 0.292 & 0.245 \\
& clDice \cite{clDice}
& 98.78 $\pm$ 0.20 & 70.12 $\pm$ 1.00 & 77.87 $\pm$ 7.11 & 23.84 $\pm$ 6.28 & 2.07 & 0.387 & 0.329 & 0.233 \\
& TopoLoss \cite{hu2019topology}
& 98.75 $\pm$ 0.19 & 71.79 $\pm$ 0.99 & 77.63 $\pm$ 7.09 & 23.65 $\pm$ 6.27 & 2.01 & 0.418 & 0.352 & 0.251 \\
& BettiLoss \cite{stucki2023topologically}
& 98.72 $\pm$ 0.20 & 69.80 $\pm$ 1.01 & 73.52 $\pm$ 6.79 & 21.65 $\pm$ 6.02 & 2.37 & 0.466 & 0.435 & 0.342 \\
& Gupta et al. \cite{gupta2024topology}
& 98.81 $\pm$ 0.19 & 72.27 $\pm$ 0.98 & 78.30 $\pm$ 7.16 & 24.61 $\pm$ 6.50 & 1.91 & 0.394 & 0.259 & 0.204 \\
& MaskVSC \cite{zhou2025masked}
& 98.83 $\pm$ 0.18 & 73.22 $\pm$ 0.97 & 79.24 $\pm$ 7.32 & 24.65 $\pm$ 6.49 & 1.78 & 0.306 & 0.223 & 0.196 \\
& \textbf{Proposed}
& \textbf{98.87 $\pm$ 0.18} & \textbf{74.48 $\pm$ 1.07} & \textbf{81.48 $\pm$ 7.51} & \textbf{26.79 $\pm$ 5.62} & \textbf{1.54} & \textbf{0.265} & \textbf{0.190} & \textbf{0.182} \\
\hline
\multirow{9}{*}{\rotatebox{90}{\textbf{F-AVSeg}}}
& TW-GAN \cite{chen2022tw}
& 98.82 $\pm$ 0.23 & 75.11 $\pm$ 1.21 & 89.32 $\pm$ 5.08 & 36.81 $\pm$ 6.19 & 0.654 & 0.609 & 0.475 & 0.315 \\
& CF-Loss \cite{zhou2024cf}
& 98.87 $\pm$ 0.22 & 76.06 $\pm$ 1.15 & 89.59 $\pm$ 4.22 & 36.48 $\pm$ 6.08 & 0.599 & 0.548 & 0.466 & 0.303 \\
& RRWNet \cite{morano2024rrwnet}
& 98.83 $\pm$ 0.19 & 76.04 $\pm$ 1.12 & 89.04 $\pm$ 3.69 & 36.12 $\pm$ 5.58 & 0.616 & 0.657 & 0.467 & 0.311 \\
& clDice \cite{clDice}
& 98.73 $\pm$ 0.24 & 75.25 $\pm$ 1.13 & 90.48 $\pm$ 2.19 & 36.46 $\pm$ 6.43 & 0.607 & 0.658 & 0.494 & 0.308 \\
& TopoLoss \cite{hu2019topology}
& 98.78 $\pm$ 0.23 & 76.05 $\pm$ 1.12 & 90.63 $\pm$ 3.22 & 36.34 $\pm$ 6.41 & 0.747 & 0.749 & 0.476 & 0.316 \\
& BettiLoss \cite{stucki2023topologically}
& 98.70 $\pm$ 0.24 & 74.76 $\pm$ 1.14 & 88.61 $\pm$ 4.27 & 36.04 $\pm$ 6.48 & 0.627 & 0.665 & 0.472 & 0.304 \\
& Gupta et al. \cite{gupta2024topology}
& 98.84 $\pm$ 0.23 & 76.81 $\pm$ 1.11 & 90.49 $\pm$ 3.14 & 37.27 $\pm$ 5.36 & 0.595 & 0.537 & 0.462 & 0.29 \\
& MaskVSC \cite{zhou2025masked}
& 98.90 $\pm$ 0.22 & 78.64 $\pm$ 1.09 & 90.78 $\pm$ 3.18 & 37.95 $\pm$ 5.43 & 0.577 & 0.524 & 0.453 & 0.285 \\
& \textbf{Proposed}
& \textbf{99.11 $\pm$ 0.21} & \textbf{81.41 $\pm$ 1.13} & \textbf{92.07 $\pm$ 2.26} & \textbf{41.91 $\pm$ 5.22} & \textbf{0.525} & \textbf{0.451} & \textbf{0.433} & \textbf{0.272} \\
\hline
\multirow{9}{*}{\rotatebox{90}{\textbf{HRF-AV}}}
& TW-GAN \cite{chen2022tw}
& 92.46 $\pm$ 0.79 & 32.19 $\pm$ 2.02 & 36.39 $\pm$ 4.78 & 3.45 $\pm$ 0.67 & 10.3 & 14.0 & 4.83 & 4.97 \\
& CF-Loss \cite{zhou2024cf}
& 92.55 $\pm$ 0.80 & 33.52 $\pm$ 1.99 & 36.40 $\pm$ 4.76 & 3.47 $\pm$ 0.65 & 9.92 & 12.3 & 4.01 & 4.61 \\
& RRWNet \cite{morano2024rrwnet}
& 92.23 $\pm$ 0.84 & 33.54 $\pm$ 2.00 & 35.98 $\pm$ 5.07 & 3.50 $\pm$ 0.66 & 10.4 & 13.5 & 4.76 & 4.84 \\
& clDice \cite{clDice}
& 91.40 $\pm$ 0.82 & 32.82 $\pm$ 2.03 & 36.35 $\pm$ 4.79 & 3.45 $\pm$ 0.70 & 10.6 & 14.1 & 4.90 & 5.01 \\
& TopoLoss \cite{hu2019topology}
& 90.77 $\pm$ 0.80 & 32.18 $\pm$ 2.02 & 35.24 $\pm$ 4.80 & 3.41 $\pm$ 0.67 & 11.2 & 14.9 & 4.73 & 5.44 \\
& BettiLoss \cite{stucki2023topologically}
& 91.82 $\pm$ 0.79 & 33.67 $\pm$ 2.06 & 36.43 $\pm$ 4.92 & 3.48 $\pm$ 0.62 & 11.5 & 12.9 & 4.51 & 4.96 \\
& Gupta et al. \cite{gupta2024topology}
& 92.40 $\pm$ 0.80 & 33.70 $\pm$ 1.99 & 37.22 $\pm$ 4.76 & 3.59 $\pm$ 0.59 & 9.86 & 12.1 & 3.94 & 4.65 \\
& MaskVSC \cite{zhou2025masked}
& 92.54 $\pm$ 0.81 & 34.25 $\pm$ 1.98 & 37.68 $\pm$ 4.74 & 3.67 $\pm$ 0.65 & 9.60 & 11.8 & 3.88 & 4.53 \\
& \textbf{Proposed}
& \textbf{92.84 $\pm$ 0.79} & \textbf{34.82 $\pm$ 1.96} & \textbf{39.04 $\pm$ 4.66} & \textbf{4.05 $\pm$ 0.61} & \textit{9.48} & \textbf{11.0} & \textbf{3.42} & \textbf{4.24} \\
\hline 
\end{tabular}
\label{compareFundus}
\end{table*}

\begin{figure*}[!ht]
\centering
\includegraphics[width=\textwidth]{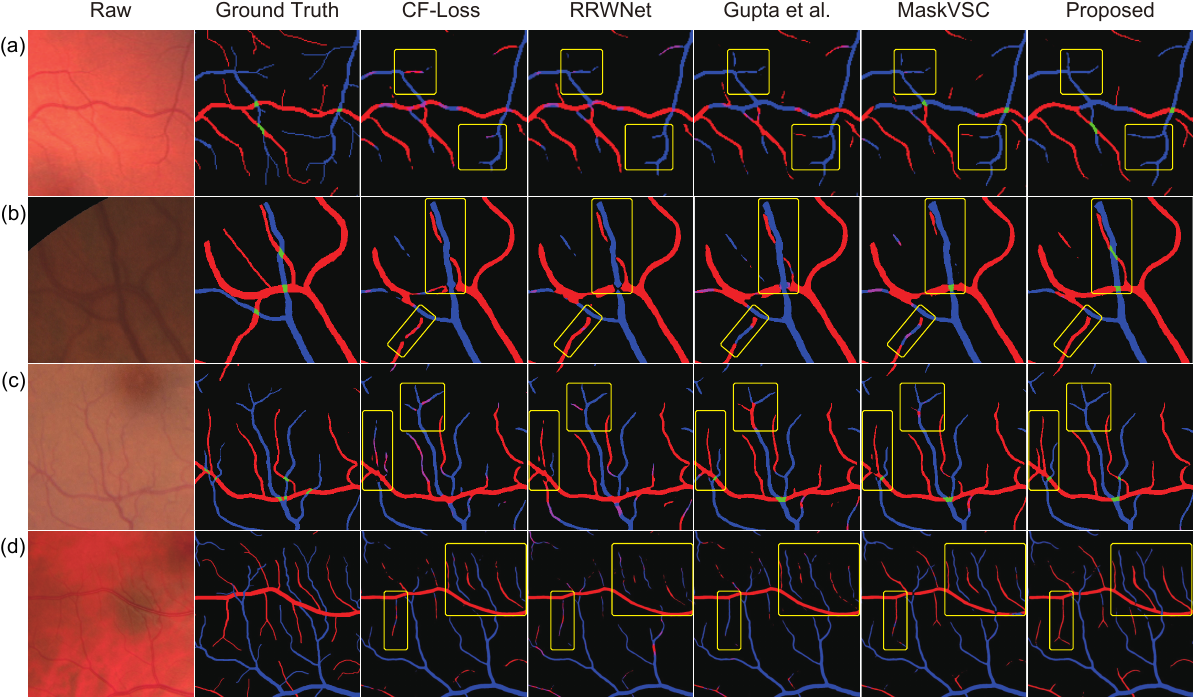}
\caption{Qualitative comparison of retinal arterial and venule tree extraction methods across four datasets: (a) RITE, (b) LES-AV, (c) F-AVSeg, (d) HRF-AV.}
\label{figcomparefundus}
\end{figure*}

\begin{table*}[!ht]
\centering
\caption{Comprehensive performance comparison on two CTA coronary artery and two TOF-MRA cerebrovascular datasets.}
\setlength{\tabcolsep}{8pt}
\begin{tabular}{cl|cccccccc}
\hline
\textbf{Dataset} & \textbf{Method} & \textbf{Dice (\%)$\uparrow$} & \textbf{clDice (\%)$\uparrow$}& \textbf{HD $\downarrow$} & $\beta_0^{err}\downarrow$ & $\beta_1^{err}\downarrow$ & $\mu_0^{err}\downarrow$ & $\mu_1^{err}\downarrow$\\
\hline
\multirow{6}{*}{\rotatebox{90}{\textbf{ImageCAS}}}
& nnU-Net \cite{isensee2021nnu}
& 78.11 $\pm$ 0.79 & 81.43 $\pm$ 5.03 & 31.49 $\pm$ 20.1 & 14.5 & 6.11 & 0.111 & 0.955 \\
& clDice \cite{clDice}
& 81.01 $\pm$ 0.71 & 83.81 $\pm$ 5.48 & 27.83 $\pm$ 21.1 & 11.4 & 4.82 & 0.085 & 0.822 \\
& cbDice \cite{cbdice}
& 80.45 $\pm$ 0.80 & 83.39 $\pm$ 5.19 & 28.53 $\pm$ 21.8 & 11.5 & 4.88 & 0.090 & 0.849 \\
& ske-recall \cite{kirchhoff2024skeleton}
& 81.68 $\pm$ 0.66 & 84.23 $\pm$ 6.08 & 27.61 $\pm$ 22.2 & 10.8 & 4.75 & 0.083 & 0.797 \\
& GLCP \cite{zhou2025glcp}
& 82.33 $\pm$ 0.58 & 85.15 $\pm$ 5.77 & 26.95 $\pm$ 21.9 & 9.97 & 4.31 & 0.076 & 0.775 \\
& \textbf{Proposed}
& \textbf{82.95 $\pm$ 0.61} & \textbf{85.95 $\pm$ 5.81} & \textit{26.58 $\pm$ 22.1} & \textbf{9.88} & \textbf{4.02} & \textit{0.073} & \textbf{0.742} \\
\hline
\multirow{6}{*}{\rotatebox{90}{\textbf{CCA}}}
& nnU-Net \cite{isensee2021nnu}
& 85.65 $\pm$ 2.44 & 86.33 $\pm$ 5.03 & 20.53 $\pm$ 7.22 & 17.7 & 10.2 & 0.305 & 0.333 \\
& clDice \cite{clDice}
& 87.62 $\pm$ 2.71 & 88.73 $\pm$ 4.81 & 17.45 $\pm$ 7.63 & 13.1 & 7.84 & 0.206 & 0.247 \\
& cbDice \cite{cbdice}
& 86.85 $\pm$ 2.96 & 88.14 $\pm$ 5.24 & 18.87 $\pm$ 8.15 & 14.8 & 8.51 & 0.222 & 0.262 \\
& ske-recall \cite{kirchhoff2024skeleton}
& 87.71 $\pm$ 2.54 & 89.64 $\pm$ 4.52 & 16.40 $\pm$ 7.61 & 12.2 & 7.37 & 0.192 & 0.233 \\
& GLCP \cite{zhou2025glcp}
& 87.98 $\pm$ 2.26 & 90.72 $\pm$ 4.17 & 14.95 $\pm$ 6.44 & 10.6 & 5.88 & 0.178 & 0.215 \\
& \textbf{Proposed}
& \textbf{88.34 $\pm$ 2.33} & \textbf{91.35 $\pm$ 4.35} & \textbf{14.22 $\pm$ 6.27} & \textbf{10.1} & \textbf{5.15} & \textbf{0.164} & \textbf{0.201} \\
\hline
\multirow{6}{*}{\rotatebox{90}{\textbf{MIDAS}}}
& nnU-Net \cite{isensee2021nnu}
& 75.51 $\pm$ 0.75 & 72.78 $\pm$ 0.37 & 12.1 $\pm$ 1.56 & 0.368 & 0.188 & 0.381 & 0.188 \\
& clDice \cite{clDice}
& 77.08 $\pm$ 0.59 & 74.25 $\pm$ 0.26 & 9.05 $\pm$ 1.85 & 0.195 & 0.131 & 0.251 & 0.141 \\
& cbDice \cite{cbdice}
& 77.92 $\pm$ 0.63 & 73.61 $\pm$ 0.45 & 10.3 $\pm$ 1.88 & 0.203 & 0.140 & 0.284 & 0.160 \\
& ske-recall \cite{kirchhoff2024skeleton}
& 78.25 $\pm$ 0.62 & 75.02 $\pm$ 0.22 & 8.33 $\pm$ 1.72 & 0.183 & 0.127 & 0.247 & 0.138 \\
& GLCP \cite{zhou2025glcp}
& 78.99 $\pm$ 0.82 & 76.28 $\pm$ 0.33 & 8.25 $\pm$ 1.82 & 0.167 & 0.109 & 0.239 & 0.122 \\
& \textbf{Proposed}
& \textbf{80.76 $\pm$ 0.70} & \textbf{77.39 $\pm$ 0.24} & \textbf{7.97 $\pm$ 1.81} & \textbf{0.155} & \textbf{0.095} & \textbf{0.233} & \textbf{0.111} \\
\hline
\multirow{6}{*}{\rotatebox{90}{\textbf{COSTA}}}
& nnU-Net \cite{isensee2021nnu}
& 93.16 $\pm$ 0.49 & 93.52 $\pm$ 1.52 & 0.993 $\pm$ 1.2 & 30.3 & 20.3 & 0.133 & 0.222 \\
& clDice \cite{clDice}
& 94.90 $\pm$ 0.63 & 95.44 $\pm$ 1.77 & 0.870 $\pm$ 1.4 & 26.1 & 17.5 & 0.101 & 0.191 \\
& cbDice \cite{cbdice}
& 94.81 $\pm$ 0.71 & 95.17 $\pm$ 1.82 & 0.931 $\pm$ 1.3 & 28.3 & 18.2 & 0.119 & 0.195 \\
& ske-recall \cite{kirchhoff2024skeleton}
& 95.77 $\pm$ 0.66 & 95.83 $\pm$ 1.68 & 0.853 $\pm$ 1.4 & 25.8 & 16.9 & 0.095 & 0.186 \\
& GLCP \cite{zhou2025glcp}
& 96.15 $\pm$ 0.61 & 96.26 $\pm$ 1.32 & 0.793 $\pm$ 1.3 & 22.4 & 16.4 & 0.088 & 0.177 \\
& \textbf{Proposed}
& \textbf{96.44 $\pm$ 0.59} & \textbf{96.67 $\pm$ 1.37} & \textbf{0.774 $\pm$ 1.3} & \textbf{20.7} & \textbf{15.9} & \textit{0.082} & \textit{0.173} \\
\hline
\end{tabular}
\label{compare3d}
\end{table*}

\subsection{Results of Comparison Experiments}

\begin{figure*}[!ht]
\centering
\includegraphics[width=\textwidth]{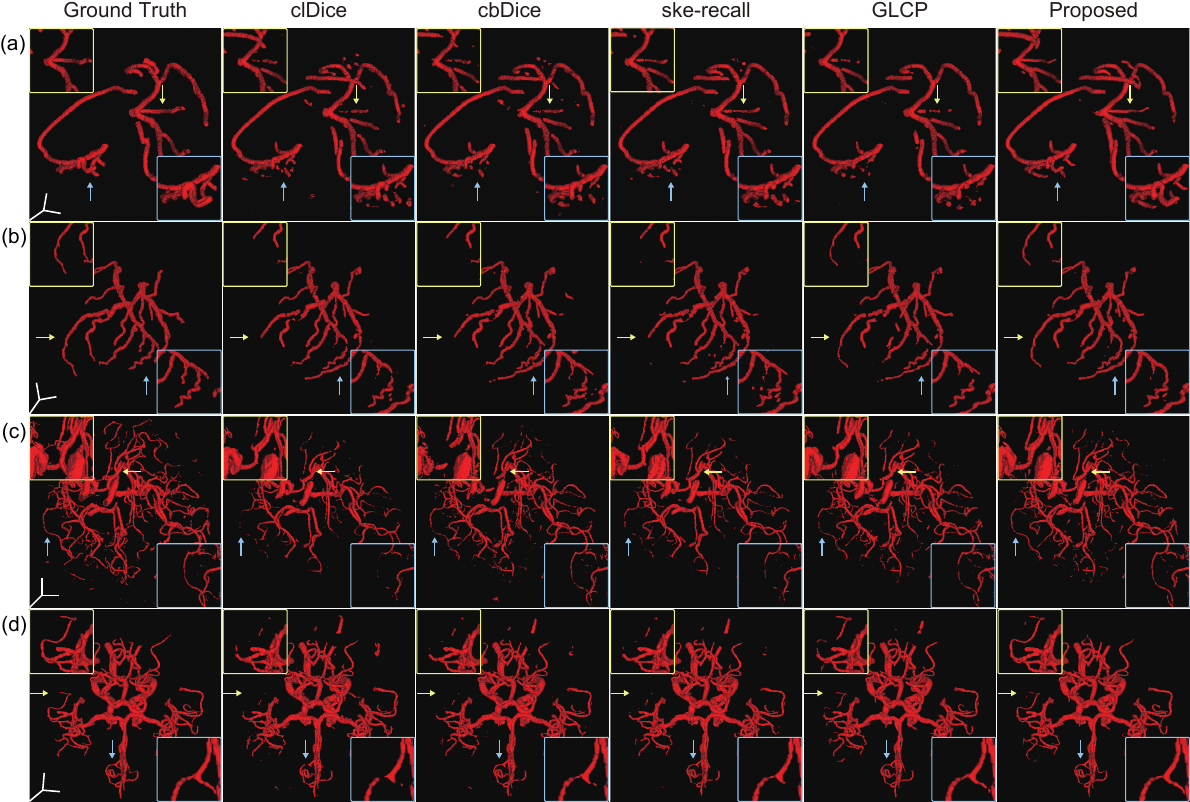}
\caption{Qualitative comparison of 3D segmentation performance on coronary artery (ImageCAS (a), CCA (b)) and cerebrovascular (MIDAS (c), COSTA (d)) datasets. Arrows indicate typical topological failures in baseline models, such as vessel fragmentation (breaks) and noisy, disconnected segments.}
\label{figcompare3d}
\end{figure*}

For 2D fundus photography, we compare MARVEL with state-of-the-art methods in three categories: (1) A/V classification and segmentation methods: \textbf{TW-GAN} \cite{chen2022tw}, \textbf{CF-Loss} \cite{zhou2024cf}, and \textbf{RRWNet} \cite{morano2024rrwnet}; (2) topology-preserving loss methods: \textbf{clDice} \cite{clDice}, \textbf{TopoLoss} \cite{hu2019topology}, and \textbf{BettiLoss} \cite{stucki2023topologically}; and (3) topology-aware methods: \textbf{Gupta et al.} \cite{gupta2024topology} and our prior \textbf{MaskVSC} \cite{zhou2025masked}. For 3D CTA and TOF-MRA, we compare MARVEL against several methods including \textbf{clDice} \cite{clDice}, \textbf{cbDice} \cite{cbdice}, \textbf{ske-recall} \cite{kirchhoff2024skeleton}, and \textbf{GLCP} \cite{zhou2025glcp}. The U-Net and nnU-Net \cite{isensee2021nnu} serve as backbones for all backbone-agnostic methods, and other methods use original settings. Statistical significance of pairwise improvements over the second-best is assessed using unpaired t-tests. In all tables \textbf{bold} denotes statistically significant improvement and \textit{italics} indicates non-significant change.

Visual examples in Fig.~\ref{figcomparefundus} and Fig.~\ref{figcompare3d} show MARVEL predictions exhibit more coherent vessel trajectories and fewer spurious branches. The improved continuity is particularly evident in capillary regions, where the model effectively leverages physiological priors to resolve ambiguous local features that pure pixel-level supervision often fails to capture.

Table~\ref{compareFundus} details the quantitative results across four 2D benchmarks. MARVEL consistently outperforms the other methods in both segmentation accuracy and topological fidelity. The most substantial gains occur on F-AVSeg, where $Dice$ increases by 2.77\% (from 78.64\% to 81.41\%) and $clDice$ improves by 1.29\% (from 90.78\% to 92.07\%). The $CAL$ metric, which quantifies vascular connectivity, demonstrates notable improvements across all datasets, particularly on RITE by 3.98\% (from 33.01\% to 36.99\%). On the challenging HRF-AV dataset with lower baseline performance, MARVEL still achieves a 1.36\% gain in $clDice$.

We also evaluate MARVEL on 3D vascular datasets. As shown in Table~\ref{compare3d}, MARVEL maintains its performance advantage across CTA and TOF-MRA datasets. The largest improvements appear on MIDAS, where $Dice$ increases by 1.77\% (from 78.99\% to 80.76\%) and $clDice$ increases by 1.11\% (from 76.28\% to 77.39\%). On the coronary artery datasets, CCA shows the most substantial reduction in $HD$ by 4.9\% (from 14.95 to 14.22). These strengths indicate that MARVEL effectively mitigates characteristic artifacts such as distal branch disconnectedness and artificial stenosis. The integration of Murray’s law provides a significant advantage in preserving the continuity of thin vessels, which often pose challenges for standard DL architectures.

Regarding topological metrics, MARVEL attains lower Betti Number errors ($\beta_0^{err}$ and $\beta_1^{err}$) across most datasets. The lower $\beta_0^{err}$ values indicate a reduction in erroneous connected components. Smaller $\beta_1^{err}$ values reflect better preservation of loops and anastomoses. A notable exception occurs with MaskVSC on the HRF-AV dataset, where its specialized connectivity loss yields competitive $\beta_0^{err}$ results. MARVEL, however, typically achieves superior $\beta_1^{err}$.

Betti Matching errors ($\mu_0^{err}$ and $\mu_1^{err}$), which penalize spatial misalignment of topological features, remain consistently low for MARVEL. These results indicate that the predicted vascular networks accurately maintain both the count and the spatial arrangement of topological elements. Overall, the integration of data-driven Murray constraints and explicit radius supervision yields measurable gains in segmentation precision, topological fidelity, and physiological plausibility.

\subsection{Performance across Different Backbones}
\begin{figure}[!ht]
\centering
\includegraphics[width=0.48\textwidth]{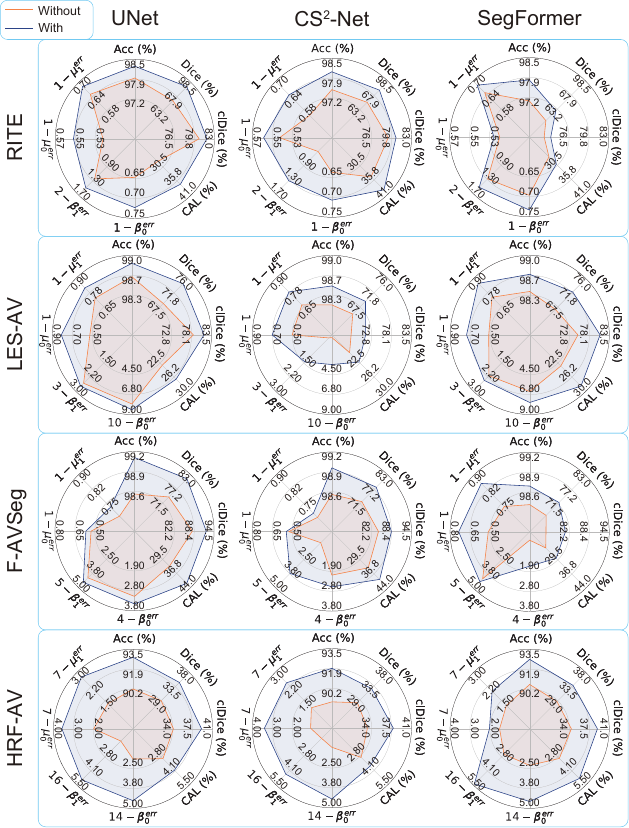}
\caption{Radar plots of backbone-agnostic evaluation of MARVEL with three different backbones on four A/V segmentation datasets. "With" and "Without" mean whether the proposed MARVEL is used.}
\label{backboneFundus}
\end{figure}

To evaluate the MARVEL’s backbone-agnostic manner, we integrate MARVEL in CNN-based models (U-Net and CS$^2$-Net \cite{mou2021cs2}), as well as a Transformer-based architecture (SegFormer \cite{xie2021segformer}). As illustrated in Fig.~\ref{backboneFundus}, MARVEL consistently enhances the performance of various base models. This backbone-agnostic characteristic suggests that our Murray's law-informed framework can serve as a regularizer for different vascular segmentation tasks. Notably, the performance gains are especially pronounced in thin-vessel regions, where architectural limitations are mitigated by the introduced biophysical priors.

\subsection{Ablation Study}

We conduct an ablation study using the U-Net backbone on the RITE dataset to assess the contribution of $\mathcal L_{Rad}$ and $\mathcal L_{Murray}$ (Table~\ref{tab4}), where $\mathcal L_{Murray}$ is further subdivided into junction probability $j(u)$ and exponent consistency $E_c$. 
The baseline U-Net provides a lower bound foundation for segmentation performance which lacks topological awareness. The addition of the $\mathcal L_{Rad}$ improves width estimation accuracy and marginally enhances segmentation metrics, while the addition of $\mathcal L_{Murray}$ (junction probability $j(u)$ and exponent consistency $E_c$) leads to substantial reduction in topological errors. The combination of both losses yields the best overall performance, confirming that radius supervision and physics-informed constraints work synergistically to guide the segmentation.

When compared to CF-Loss \cite{zhou2024cf}, a loss specifically designed for artery and vein classification, $\mathcal L_{Rad}$ attains higher $clDice$, $\beta_1^{err}$, and $\mu^{err}$, although CF-Loss remains competitive on $Acc$, $Dice$, and $\beta_0^{err}$. Relative to other backbone-agnostic topology preserving objectives (clDice \cite{clDice}, TopoLoss \cite{hu2019topology}, BettiLoss \cite{stucki2023topologically}), $\mathcal L_{Rad}$ improves both classification and segmentation. Adding $\mathcal L_{Murray}$ increases margins over these alternatives in $Acc$, $Dice$, $clDice$ and reduces $\beta^{err}$ and $\mu^{err}$. In combination, the two losses yield statistically significant improvements across the evaluated measures (Table~\ref{compareFundus}).

\begin{table}[!t]
\centering
\scriptsize
\caption{Ablation study on MARVEL's components using U-Net backbone on the RITE dataset. $\checkmark$ stands for "included". $\mathcal L_{R}$, $j(u)$, and $E_c$ mean $\mathcal L_{Rad}$, junction probability, and exponent consistency, respectively. The $j(u)$ and $E_c$ constitute $\mathcal L_{Murray}$.}
\setlength{\tabcolsep}{3pt}
\begin{tabular}{ccc|cccccc}
\hline
$\mathcal L_{R}$ & $j(u)$ & $E_c$ & \textbf{Dice} (\%)$\uparrow$ & \textbf{clDice} (\%)$\uparrow$ & $\beta_0^{err}\downarrow$ & $\beta_1^{err}\downarrow$ & $\mu_0^{err}\downarrow$ & $\mu_1^{err}\downarrow$\\
\hline
& & 
& 66.52 & 80.51 & 0.353 & 0.760 & 0.472 & 0.350 \\
$\checkmark$ & & 
& 68.64 & 81.37 & 0.315 & 0.607 & 0.473 & 0.344 \\
& & $\checkmark$ 
& 68.30 & 81.08 & 0.337 & 0.652 & 0.470 & 0.346 \\
$\checkmark$ & & $\checkmark$ 
& 68.95 & 81.43 & 0.301 & 0.557 & 0.467 & 0.338 \\
& $\checkmark$ & $\checkmark$ 
& 69.64 & 81.61 & 0.295 & 0.560 & 0.467 & 0.340 \\
$\checkmark$ & $\checkmark$ & $\checkmark$
& \textbf{71.12} & \textbf{81.78} & \textbf{0.277} & \textbf{0.505} & \textbf{0.446} & \textbf{0.317} \\
\hline
\end{tabular}
\label{tab4}
\end{table}

\begin{figure}[!ht]
\centering
\includegraphics[width=0.48\textwidth]{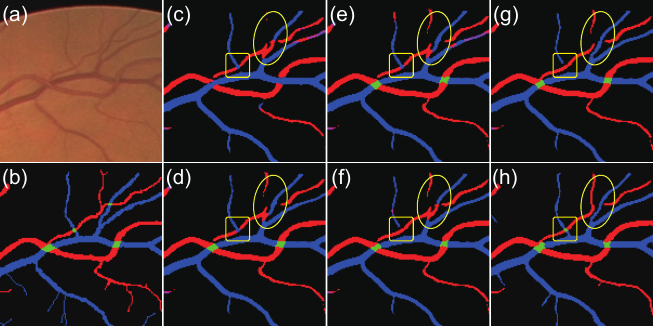}
\caption{Ablation study visualization on RITE dataset. (a) Raw image, (b) Ground truth, (c) U-Net, (d) U-Net + $\mathcal L_{Rad}$, (e) U-Net + exponent consistency, (f) U-Net + $\mathcal L_{Rad}$ + exponent consistency, (g) U-Net + $\mathcal L_{Murray}$, (h) U-Net + $\mathcal L_{Rad}$ + $\mathcal L_{Murray}$ (Proposed MARVEL).}
\label{fig6}
\end{figure}

\subsection{Performance under Different Resolution Priors}
We investigate whether the Murray exponent mapping is best derived from a high-resolution reference or it can be transferred from low-resolution datasets. As reported in Table~\ref{tab5}, we compare the performance of MARVEL when guided by the high-resolution priors versus other low-resolution datasets. While both approaches yield consistent improvements, using a high-resolution dataset as priors yields further gains across all metrics. This suggests that low-resolution datasets contain measurement noise that corrupts the empirical Murray exponent distribution. By leveraging more accurate physical relationships observed in high-fidelity data, we can effectively transfer this precise hemodynamic knowledge to lower-quality domains. 

\begin{table*}[!ht]
\centering
\caption{Impact of priors on segmentation performance. "None" means without guidance of any dataset but with the radius map loss $\mathcal L_{Rad}$.}
\begin{tabular}{ll|cccccccc}
\hline
\textbf{Dataset} & \textbf{Priors Dataset} & \textbf{Acc (\%)$\uparrow$} & \textbf{Dice (\%)$\uparrow$} & \textbf{clDice (\%)$\uparrow$}& \textbf{CAL (\%)$\uparrow$} & $\beta_0^{err}\downarrow$ & $\beta_1^{err}\downarrow$ & $\mu_0^{err}\downarrow$ & $\mu_1^{err}\downarrow$\\
\hline
\multirow{3}{*}{\textbf{RITE}}
& None
& 98.15 $\pm$ 0.21 & 68.64 $\pm$ 1.03 & 81.37 $\pm$ 2.90 & 31.79 $\pm$ 3.16 & 0.315 & 0.607 & 0.470 & 0.344 \\
& Low-resolution
& 98.18 $\pm$ 0.20 & 69.25 $\pm$ 0.93 & 81.44 $\pm$ 2.68 & 32.08 $\pm$ 3.48 & 0.298 & 0.554 & 0.467 & 0.340 \\
& High-resolution
& \textbf{98.28 $\pm$ 0.18} & \textbf{71.12 $\pm$ 0.92} & \textbf{81.78 $\pm$ 2.65} & \textbf{36.99 $\pm$ 3.67} & \textbf{0.277} & \textbf{0.505} & \textbf{0.446} & \textbf{0.317} \\
\hline
\multirow{3}{*}{\textbf{LES-AV}}
& None
& 98.78 $\pm$ 0.19 & 73.02 $\pm$ 0.87 & 78.87 $\pm$ 7.52 & 24.25 $\pm$ 6.30 & 1.91 & 0.333 & 0.236 & 0.203 \\
& Low-resolution
& 98.82 $\pm$ 0.17 & 73.40 $\pm$ 0.96 & 79.33 $\pm$ 7.65 & 24.72 $\pm$ 6.17 & 1.76 & 0.302 & 0.211 & 0.192 \\
& High-resolution
& \textbf{98.87 $\pm$ 0.18} & \textbf{74.48 $\pm$ 1.07} & \textbf{81.48 $\pm$ 7.51} & \textbf{26.79 $\pm$ 5.62} & \textbf{1.54} & \textbf{0.265} & \textbf{0.190} & \textbf{0.182} \\
\hline
\multirow{3}{*}{\textbf{F-AVSeg}}
& None
& 98.73 $\pm$ 0.21 & 77.65 $\pm$ 1.08 & 89.92 $\pm$ 2.44 & 36.75 $\pm$ 5.74 & 0.614 & 0.588 & 0.465 & 0.290 \\
& Low-resolution
& 98.89 $\pm$ 0.20 & 78.82 $\pm$ 1.05 & 90.83 $\pm$ 2.65 & 38.02 $\pm$ 5.60 & 0.572 & 0.511 & 0.451 & 0.286 \\
& High-resolution
& \textbf{99.11 $\pm$ 0.21} & \textbf{81.41 $\pm$ 1.13} & \textbf{92.07 $\pm$ 2.26} & \textbf{41.91 $\pm$ 5.22} & \textbf{0.525} & \textbf{0.451} & \textbf{0.433} & \textbf{0.272} \\
\hline
\end{tabular}
\label{tab5}
\end{table*}

\begin{figure*}[!ht]
\centering
\includegraphics[width=\textwidth]{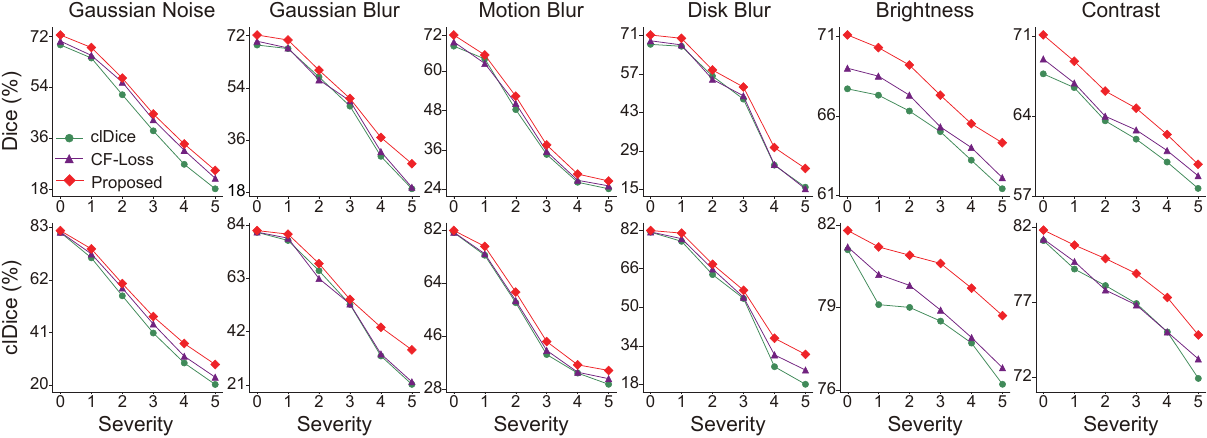}
\caption{Robustness evaluation on corrupted images. Performance degradation curves of Dice and clDice metrics for clDice \cite{clDice}, CF-Loss \cite{zhou2024cf}, and Proposed method on six different corruption types increasingly degraded RITE images.}
\label{robust}
\end{figure*}

\subsection{Robustness on Low-quality Fundus Images}
To assess the resilience of MARVEL in real-world clinical environments characterized by suboptimal images, we synthetically apply six distinct corruption protocols on RITE test images, including Gaussian noise, Gaussian blur, motion blur, disk blur, and variations in brightness and contrast. We compare clDice \cite{clDice}, CF-Loss \cite{zhou2024cf}, and the proposed MARVEL using a U-Net backbone on the RITE dataset. Fig.~\ref{robust} illustrates the $Dice$ scores and $clDice$ metrics for each corruption type across severity levels.

MARVEL consistently outperforms clDice and CF-Loss across all corruption types and severity levels. This robust advantage is evident in both segmentation accuracy ($Dice$) and the preservation of topological structures ($clDice$). For specific corruptions such as Gaussian blur, disk blur, and brightness variations, the performance gap becomes more pronounced at higher severity levels. Specifically, for Gaussian blur at severity level 5, the proposed method achieves an 8\% improvement in $Dice$ and a 12.6\% gain in $clDice$. These findings demonstrate that incorporating Murray-based priors enhance model resilience and enables the generation of biologically plausible segmentations even when the visual signal is compromised.

\begin{figure}[!ht]
\centering
\includegraphics[width=0.48\textwidth]{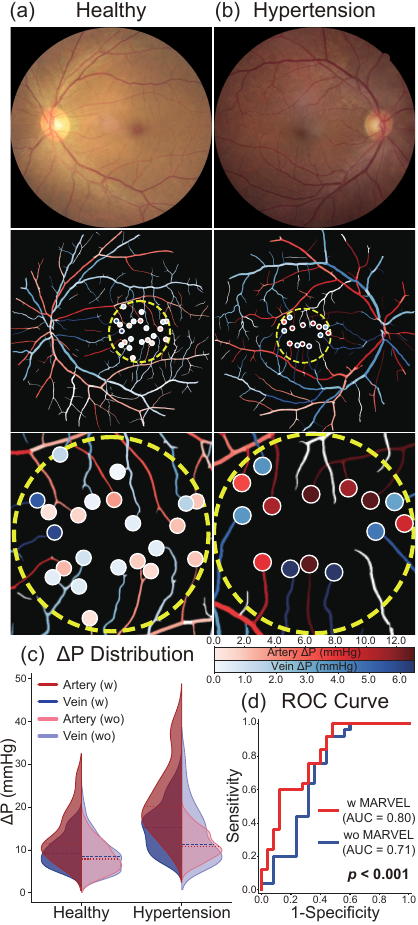}
\caption{Visualization of blood pressure difference in vessels of healthy (a) and hypertension (b) eyes, with the macular region indicated by the yellow dashed line. (c) Distribution of blood pressure difference in healthy and hypertensive eyes for both models, with dash lines representing the mean values. (d) ROC curve and AUC score for both models of hypertension diagnosis.}
\label{downstream}
\end{figure}

\subsection{Application in Hypertension Classification}
Hypertension induces systemic microvascular remodeling, characterized by arteriolar narrowing and rarefaction \cite{cheung2012retinal}. To validate the hemodynamic fidelity of MARVEL, we quantify the arteriovenous blood pressure difference in the macula as a biomarker for hypertension. We employ a U-Net backbone to segment retinal vessels from 25 healthy and 25 hypertensive fundus images, acquired from the Singapore National Eye Centre and de-identified. The hemodynamic simulation pipeline is detailed below.

\subsubsection{Graph-Based Vascular Representation}
The hemodynamic characteristics of the retinal vasculature are modeled via a graph-based fluid simulation. We first refine the predicted arterial and venous masks via morphological closing and small-region removal. Subsequently, we extract topological skeletons from the refined masks and convert them into a directed graph $G(V, E)$, where each edge represents a vessel segment characterized by its pixel length $L$ and radius $r$.

\subsubsection{Boundary Conditions and Pressure Initialization}
To initialize the fluid simulation, specific boundary conditions are established at the optic disc. We assign an inlet retinal arterial pressure of 76.9 mm Hg \cite{jonas2004ophthalmodynamometric} and an outlet retinal venous pressure of 21 mm Hg \cite{flammer2015retinal}. These benchmarks establish a standardized computational framework, allowing for the comparison of relative pressure gradients across subjects despite systemic physiological variations. 

\subsubsection{Adaptive Murray’s Law for Flow Distribution}
The distribution of blood flow $Q$ at each vascular bifurcation is governed by the local vascular geometry. Unlike standard models that assume a fixed cubic relationship, we derive the flow partitioning from the adaptive exponent $\alpha$. The flow $Q$ is partitioned according to the following relationship:
\begin{equation}
Q_{i} = Q_{0} \cdot \left( \frac{r_{i}^\alpha}{r_1^\alpha + r_2^\alpha} \right).
\end{equation}

By incorporating the adaptive exponent, this formulation implicitly captures the structural remodeling and non-Newtonian flow characteristics relevant to hypertensive pathology. 

\subsubsection{Pressure Calculation and Hydraulic Resistance}
Following the determination of flow rates, pressure values are computed iteratively starting from the optic disc. The pressure difference $\Delta P$ across each segment is modeled using Poiseuille’s Law:
\begin{equation}
\Delta P = Q \times R = K \times ||Q|| \times \left( \frac{8\eta L}{\pi r^4} \right),
\end{equation}
where $R$ denotes hydraulic resistance, $K$ is a scaling constant used to calibrate pixel-based measurements into simulation pressure units, $||Q||$ is unit flow, $\eta$ represents blood viscosity, $L$ is the pixel length, and $r$ is the radius. Measurement of Q requires a dual-beam Doppler OCT system which is not commonly equipped in clinics \cite{luft2016ocular,luft2016measurements}. The breadth-first search algorithm \cite{moore1959shortest} is used to trace vascular paths from the optic disc toward the macula. The arteriovenous blood pressure difference $\Delta P_{AV}$ is determined by subtracting the cumulative pressure difference along the arterial and venous paths:
\begin{equation}
\begin{aligned}
\Delta P_{AV} &= \Delta P_{in \to out} - (\Delta P_{A} - \Delta P_{V})\\
&= \Delta P_{in \to out} - (\sum_{i \in Apath} \Delta P_i - \sum_{j \in Vpath} \Delta P_j).
\end{aligned}
\end{equation}

\subsubsection{Hypertension Classification}
We calculate one $\Delta P_{AV}$ value per eye within the macular region as a classification feature for hypertension. Fig.~\ref{downstream} (a) and (b) visualize the steeper pressure gradients in hypertensive eyes than in healthy controls, where darker colors indicate steeper pressure gradients in vessel segments and higher cumulative pressure differences from optic disc to the macular endpoints. This aligns with established pathophysiology: the narrowing of arterioles increases hydraulic resistance, causing a more rapid dissipation of pressure before the capillary bed. 

Fig.~\ref{downstream} (c) illustrates the distribution of arterial-venous pressure differences in healthy and hypertensive eyes, where darker and lighter shades represent models with and without MARVEL respectively. $\Delta P_{AV}$ in hypertensive eyes is greater than in healthy eyes, which is consistent with clinical observations \cite{fukutsu2021deep}. Moreover, MARVEL can yield a more distinct distribution than the baseline. Classification performance is assessed using the Receiver Operating Characteristic (ROC) curve and Area Under the Curve (AUC) score shown in Fig.~\ref{downstream} (d). The MARVEL-integrated model achieves superior separation of $\Delta P_{AV}$ between the healthy and hypertensive eyes compared to the baseline (DeLong test \cite{delong1988comparing}, $p < 0.001$). This confirms that MARVEL achieves enhanced classification accuracy and preserves pathological vascular details essential for hemodynamic modeling.

To further characterize morphological differences, we compute three vascular topology metrics: branch angle, defined as the supplement of the angle between parent and child vessel vectors; radius continuity, measured as the standard deviation of vessel radii along paths between endpoints and bifurcations; and bifurcation asymmetry, defined as the normalized ratio of child branch radii at each junction. Table~\ref{tab6} reports the distribution of each metric for vascular masks obtained with and without MARVEL, stratified by clinical group, with 95\% confidence intervals. Across all conditions, MARVEL consistently yields more complete and physiologically plausible retinal vascular networks.

\begin{table*}[!ht]
\centering
\caption{Comparison of quantitative metrics for healthy and hypertensive eyes. $\angle$, $RC$, and $BA$ mean branch angle, radius continuity, and bifurcation asymmetry, and the subscript A/V mean artery and vein. Mean values and 95\% confidence intervals are reported.}
\begin{tabular}{ll|ccc|ccc}
\hline
\textbf{MARVEL} & \textbf{Condition} & \textbf{$\angle_A$ ($\degree$)} & \textbf{$RC_A$ $\downarrow$} & \textbf{$BA_A$ $\downarrow$}& \textbf{$\angle_V$ ($\degree$)} & \textbf{$RC_V$ $\downarrow$} & \textbf{$BA_V$ $\downarrow$}\\
\hline
\multirow{2}{*}{\textbf{Without}}
& Healthy
& 60.41 \scriptsize{(59.65, 61.17)} & 1.10 \scriptsize{(1.06, 1.15)} & 0.24 \scriptsize{(0.23, 0.25)} & 59.62 \scriptsize{(59.05, 60.20)} & 1.04 \scriptsize{(0.98, 1.09)} & 0.22 \scriptsize{(0.21, 0.23)} \\
& Hypertension
& 57.20 \scriptsize{(56.35, 58.05)} & 1.00 \scriptsize{(0.97, 1.03)} & 0.23 \scriptsize{(0.22, 0.25)} & 58.06 \scriptsize{(57.37, 58.75)} & 0.95 \scriptsize{(0.91, 1.00)} & 0.23 \scriptsize{(0.22, 0.24)} \\
\hline
\multirow{2}{*}{\textbf{With}}
& Healthy
& 55.07 \scriptsize{(54.27, 55.86)} & 1.00 \scriptsize{(0.95, 1.05)} & 0.18 \scriptsize{(0.17, 0.18)} & 55.86 \scriptsize{(55.23, 56.50)} & 1.03 \scriptsize{(0.97, 1.08)} & 0.17 \scriptsize{(0.16, 0.18)} \\
& Hypertension
& 54.48 \scriptsize{(53.72, 55.23)} & 0.89 \scriptsize{(0.87, 0.92)} & 0.19 \scriptsize{(0.18, 0.20)} & 55.89 \scriptsize{(55.24, 56.53)} & 0.94 \scriptsize{(0.91, 0.97)} & 0.17 \scriptsize{(0.16, 0.18)} \\
\hline
\end{tabular}
\label{tab6}
\end{table*}

\section{Conclusion}
In this work, we addressed the fundamental disconnect between data-driven segmentation and the biophysical laws governing vascular physiology. We introduced MARVEL, a framework that embeds Murray’s Law as a differentiable constraint, ensuring that the segmentation outputs achieve both pixel-level accuracy and hemodynamic validity. Unlike methods that rely on rigid assumptions, MARVEL utilizes an adaptive width-exponent mapping to enforce biologically accurate branching constraints across diverse vascular beds.

Our evaluation on eight datasets confirms that this physics-informed approach yields superior topological consistency across retinal, coronary, and cerebrovascular imaging modalities. Our clinical validation on hypertensive cohorts demonstrates that these topological improvements translate directly into diagnostic power for hypertension: by preserving the fine-scale connectivity required for hemodynamic simulation, MARVEL enabled the precise calculation of arteriovenous pressure gradients, significantly outperforming baseline models in hypertension classification.

Future research could explore several directions. Since pathological features like intraretinal microvascular abnormalities and neovascularization often violate optimal transport principles, future iterations of MARVEL could incorporate adaptive regularization mechanisms that learn to relax physical constraints in anomalous regions. Another direction involves scaling this hemodynamic validation from 2D retinal models to full 3D volumetric simulations that include the capillary beds.

\bibliographystyle{IEEEtran}
\bibliography{references}

\vspace{11pt}

\begin{IEEEbiography}[{\includegraphics[width=1in,height=1.25in,clip,keepaspectratio]{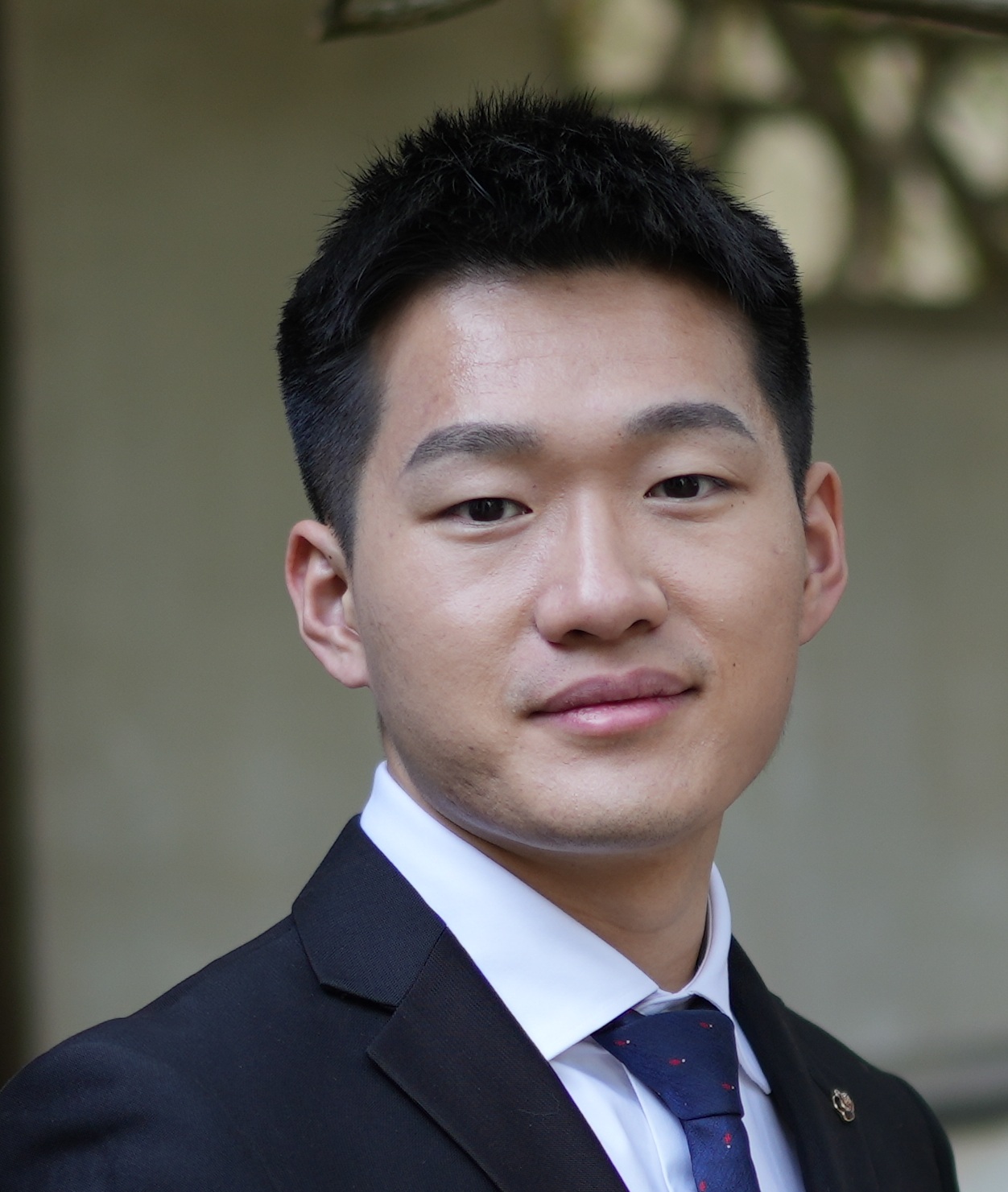}}]{Yi Zhou} is a Research Fellow at Singapore Eye Research Institute, Singapore National Eye Centre. He received his B. E. degree and Ph.D. degree from Soochow University in 2018 and 2023, advised by Prof. Xinjian Chen. His research interests include, but not limited to, medical image analysis, image generation, and AI in ophthalmology. He received Travel Award at MICCAI 2021 and ARVO 2026. He is an Associate Editor of Medical Physics.
\end{IEEEbiography}

\begin{IEEEbiography}[{\includegraphics[width=1in,height=1.25in,clip,keepaspectratio]{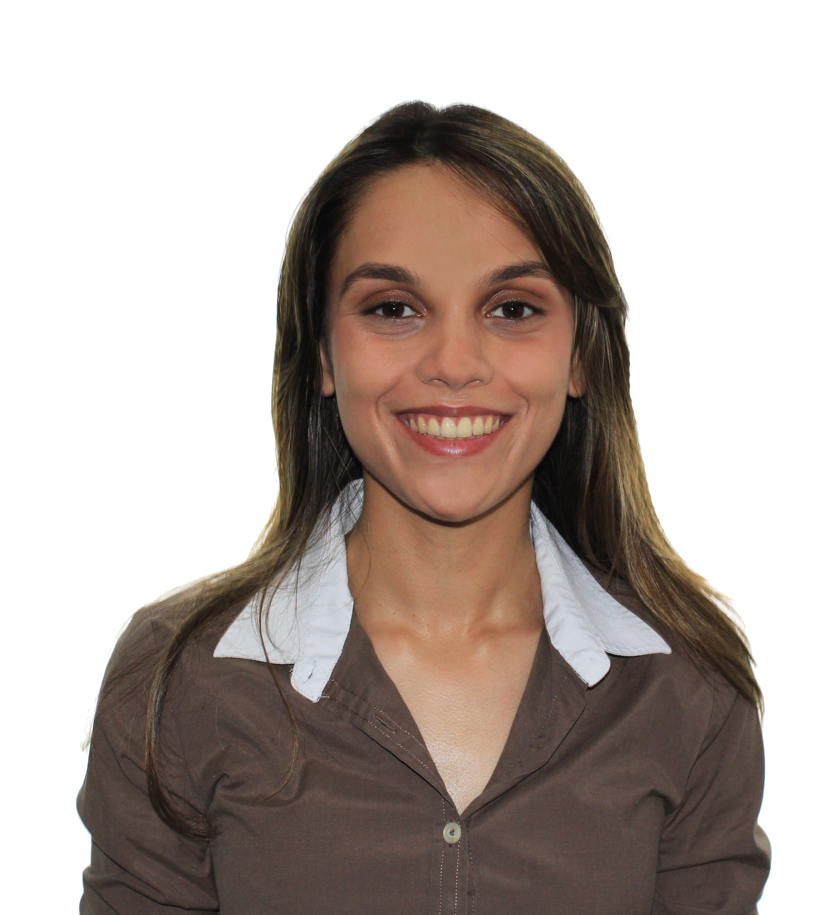}}]{Thiara Sana Ahmed} is now a Ph.D. student at the school of chemistry, chemical engineering, and biotechnology at Nanyang Technological University, Singapore. Advised by Prof. Leopold Schmetterer. Her research interests include, but not limited to, temporal disease modeling, and machine learning for medical imaging.
\end{IEEEbiography}

\begin{IEEEbiography}[{\includegraphics[width=1in,height=1.25in,clip,keepaspectratio]{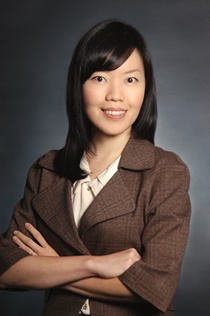}}]{Jacqueline Chua} is an Associate Professor at Duke–NUS Medical School, Singapore, and Deputy Head and Clinician Scientist in the Ocular Imaging Group at the Singapore Eye Research Institute, Singapore National Eye Centre. She is an optometrist and clinician-scientist whose research focuses on developing imaging biomarkers for early detection and monitoring of eye and systemic diseases. Her work integrates optical coherence tomography (OCT/OCTA), retinal image analysis, and artificial intelligence to improve disease detection, image quality, and clinical decision support.
\end{IEEEbiography}

\begin{IEEEbiography}[{\includegraphics[width=1in,height=1.25in,clip,keepaspectratio]{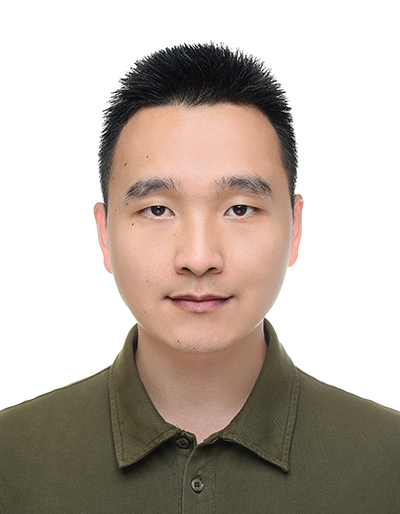}}]{Meng Wang} is a Senior Research Fellow at the Centre for Innovation and Precision Eye Health and the department of Ophthalmology, Yong Loo Lin School of Medicine, National University of Singapore. His research focuses on artificial intelligence algorithm development, with an emphasis on computer vision, vision-language foundation models, and trustworthy AI for healthcare.
\end{IEEEbiography}

\begin{IEEEbiography}[{\includegraphics[width=1in,height=1.25in,clip,keepaspectratio]{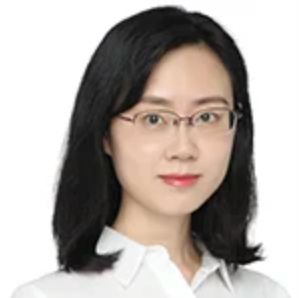}}]{Qinrong Zhang} is an Assistant Professor in Biomedical Engineering at City University of Hong Kong. Her research centers on the development of advanced optical microscopy and computational imaging technologies for biological applications, with an emphasis on in vivo imaging of the brain and eye. 
\end{IEEEbiography}

\begin{IEEEbiography}[{\includegraphics[width=1in,height=1.25in,clip,keepaspectratio]{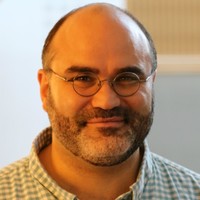}}]{Alejandro F. Frangi} (Fellow, IEEE) is the Bicentenary Turing Chair in Computational Medicine at the University of Manchester, Manchester, U.K., where he holds joint appointments in the Schools of Engineering and Health Sciences. He is also the Royal Academy of Engineering Chair in Emerging Technologies. He earned his Ph.D. degree in medicine from the University Medical Centre Utrecht in 2001. His research focuses on precision computational medicine, \textit{in silico} medical device applications, and medical image computing. He is an Associate Editor for the \textsc{IEEE Transactions on Pattern Analysis and Machine Intelligence} and the \textsc{IEEE Transactions on Medical Imaging}.
\end{IEEEbiography}

\begin{IEEEbiography}[{\includegraphics[width=1in,height=1.25in,clip,keepaspectratio]{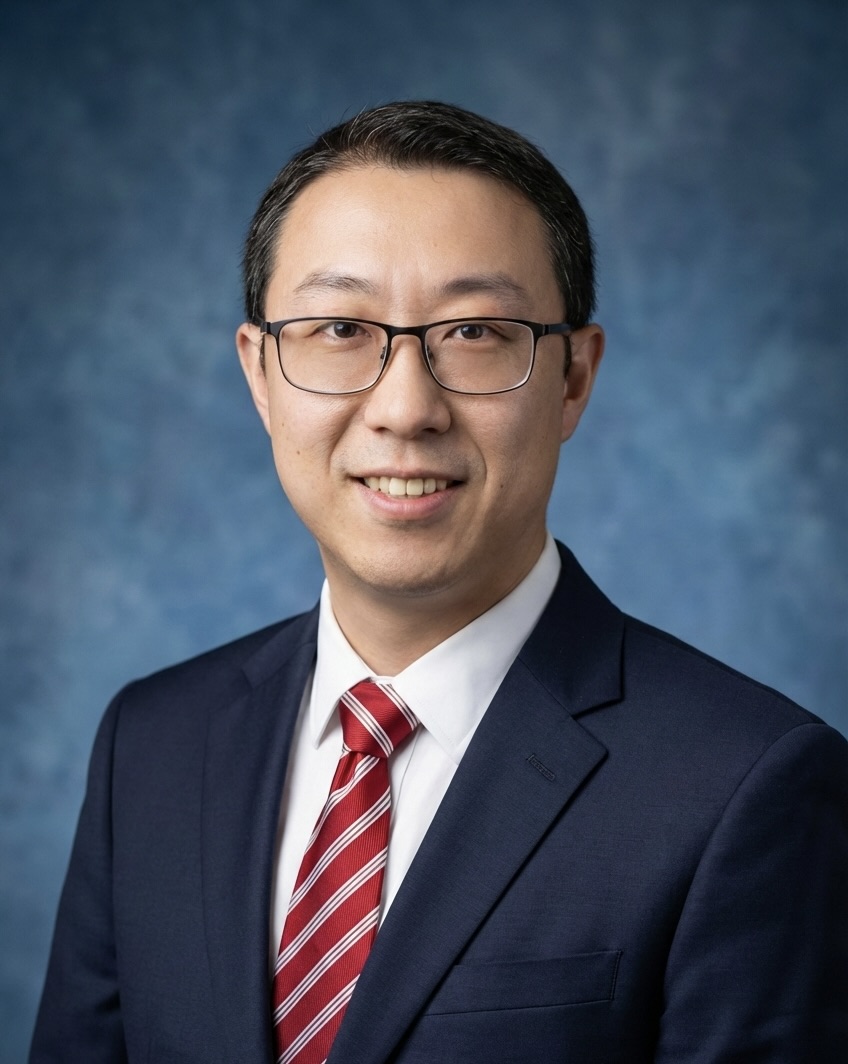}}]{Huazhu Fu} (Senior Member, IEEE) is a Principal Scientist at the Institute of High Performance Computing (IHPC), A*STAR, Singapore. He earned his Ph.D. from Tianjin University in 2013. His research focuses on medical image analysis, AI for healthcare, and trustworthy AI. He is an Associate Editor for several distinguished journals, including \textsc{IEEE Transactions on Medical Imaging}, \textsc{IEEE Transactions on Neural Networks and Learning Systems}, and \textsc{IEEE Journal of Biomedical and Health Informatics}.
\end{IEEEbiography}

\begin{IEEEbiography}[{\includegraphics[width=1in,height=1.25in,clip,keepaspectratio]{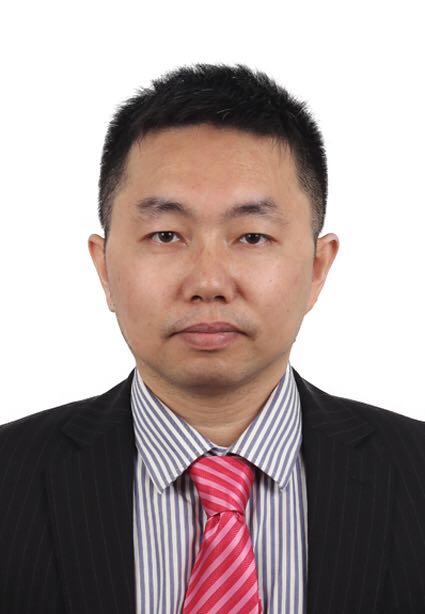}}]{Jun Cheng} (Senior Member, IEEE) received the B.E. degree in electronic engineering and information science from the University of Science and Technology of China, and the Ph.D. degree from Nanyang Technological University, Singapore. He is currently a Principal Research Scientist with the Institute for Infocomm Research, A*STAR, working on AI for medical imaging, robust machine vision, and perception. Dr. Cheng is an Associate Editor of \textsc{IEEE Transactions on Medical Imaging} and a Senior Area Editor of \textsc{IEEE Transactions on Image Processing}.
\end{IEEEbiography}

\begin{IEEEbiography}[{\includegraphics[width=1in,height=1.25in,clip,keepaspectratio]{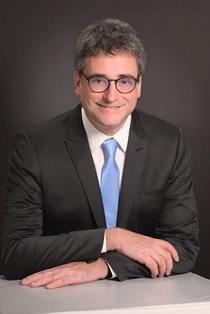}}]{Leopold Schmetterer} is head of Ocular Imaging and Senior Scientific Director at Singapore Eye Research Institute (SERI). He is a specialist in ocular imaging and has an interest in vascular and neurodegenerative eye disease. His major aim is to translate imaging technology from bench towards clinical applications.
\end{IEEEbiography}

\begin{IEEEbiography}[{\includegraphics[width=1in,height=1.25in,clip,keepaspectratio]{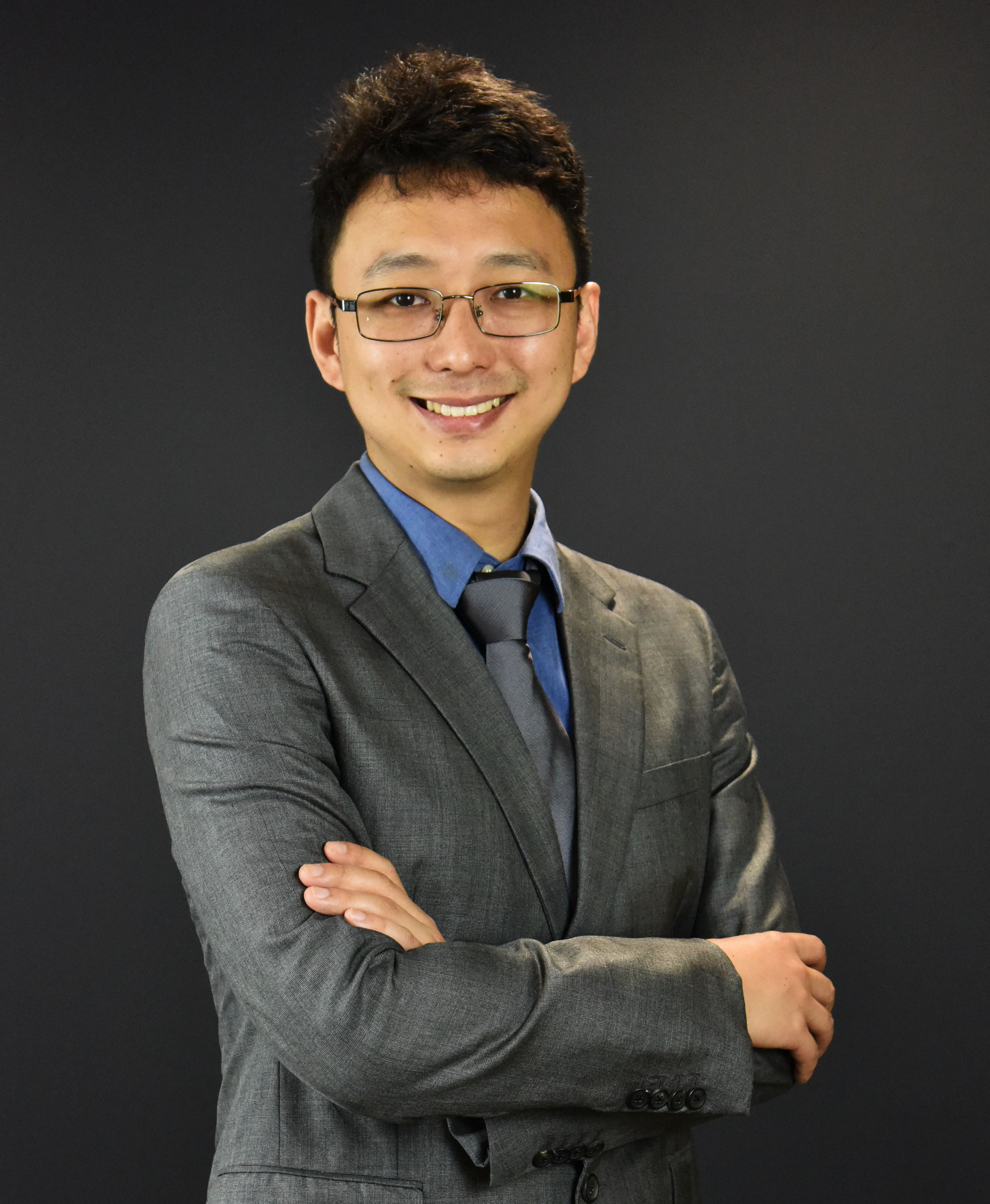}}]{Bingyao Tan} received his PhD in physics from University of Waterloo, Canada. He is a physicist whose research focuses on the development and clinical translation of advanced ophthalmic imaging technologies, particularly optical coherence tomography (OCT). He is currently a Junior Principal Investigator at the Singapore Eye Research Institute (SERI) and an Assistant Professor at Duke–NUS Medical School. His laboratory integrates optical engineering, computational algorithms, and clinical studies to address problems of high clinical relevance in medicine.
\end{IEEEbiography}

\vfill

\end{document}